\documentclass[twoside]{article}

\usepackage[accepted]{aistats2022_arxiv}

\usepackage[round]{natbib}
\renewcommand{\bibname}{References}

\usepackage[utf8]{inputenc} % allow utf-8 input
\usepackage[T1]{fontenc}    % use 8-bit T1 fonts
\usepackage{hyperref}       % hyperlinks
\hypersetup{colorlinks=true, citecolor=blue, linkcolor=blue}
\usepackage{url}            % simple URL typesetting
\usepackage{booktabs}       % professional-quality tables
\usepackage{amsfonts}       % blackboard math symbols
\usepackage{nicefrac}       % compact symbols for 1/2, etc.
\usepackage{microtype}      % microtypography
\usepackage{xcolor}         % colors

\usepackage{amsfonts, graphicx,color,multirow, amsthm,amsmath}
\usepackage{setspace,url}
\RequirePackage{amsgen,amsmath,amstext,amsbsy,amsopn,amssymb,amscd,graphicx,cancel,needspace,mathtools,bm}
\usepackage{multirow}
\usepackage{graphicx}
\usepackage{bm}
\usepackage[inline,shortlabels]{enumitem}
\usepackage{amsthm, amsmath, amssymb}
\usepackage[toc,page]{appendix}
\usepackage{caption}
\usepackage{subcaption}
\usepackage{todonotes}

\usepackage{wrapfig,lipsum,booktabs}

\usepackage{neb}

\begin{document}

\twocolumn[

\aistatstitle{Nonparametric Empirical Bayes Estimation and Testing for Sparse and Heteroscedastic Signals}

\aistatsauthor{ Junhui Cai \And Xu Han \And  Ya'acov Ritov \And Linda Zhao }

\aistatsaddress{ University of Pennsylvania \And Temple University \And University of Michigan \And University of Pennsylvania } ]

% \maketitle

%%%%%%%%%%%%%%%%%%%%%%%%%%%%%%%%%%%%%%%%%%%%%%%%%%%%%%%%%%%%
\begin{abstract}
% The simultaneous estimation and testing of high-dimensional parameters has been a key problem for large-scale modern data. The goal is to identify the sparse signals, ``the needles in the haystack'', and to estimate the signal strength. However, the unprecedented complexity and heterogeneity in modern data structure present a significant statistical challenge -- to effectively exploit commonalities and to robustly adjust for both sparsity and heterogeneity. In this paper, we propose a Spike-and-Nonparametric prior (SNP), by cross-fertilizing the widely adopted spike-and-slab paradigm in Bayesian variable selection and a flexible non-parametric structure. We adopt the empirical Bayes methodology to estimate the prior and produce the entire posterior distribution. With the posterior distribution, we are able to produce point estimate with shrinkage or soft-thresholding property and uncertainty quantification. We further propose an optimal multiple testing procedure. Our method exhibits promising empirical performance on both simulated data and a real gene expression data.

Large-scale modern data often involves estimation and testing for high-dimensional unknown parameters. It is desirable to identify the sparse signals, ``the needles in the haystack'', with accuracy and false discovery control.  However, the unprecedented complexity and heterogeneity in modern data structure require new machine learning tools to effectively exploit commonalities and to robustly adjust for both sparsity and heterogeneity. In addition, estimates for high-dimensional parameters often lack uncertainty quantification. In this paper, we propose a novel Spike-and-Nonparametric mixture prior (SNP) -- a spike to promote the sparsity and a nonparametric structure to capture signals. 
We adopt the empirical Bayes methodology and
% In contrast to the state-of-the-art methods, the proposed methods 
solve the estimation and testing problem at once with several merits: 1) an accurate sparsity estimation; 2) point estimates with shrinkage/soft-thresholding property; 3) credible intervals for uncertainty quantification; 4) an optimal multiple testing procedure that controls false discovery rate. Our method exhibits promising empirical performance on both simulated data and a gene expression case study.
\end{abstract}
%%%%%%%%%%%%%%%%%%%%%%%%%%%%%%%%%%%%%%%%%%%%%%%%%%%%%%%%%%%%

%%%%%%%%%%%%%%%%%%%%%%%%%%%%%%%%%%%%%%%%%%%%%%%%%%%%%%%%%%%%
%%%%%%%%%%%%%%%%%%%%%%%%%%%%%%%%%%%%%%%%%%%%%%%%%%%%%%%%%%%%
\section{Introduction}
\label{sec:intro}
%%%%%%%%%%%%%%%%%%%%%%%%%%%%%%%%%%%%%%%%%%%%%%%%%%%%%%%%%%%%
%%%%%%%%%%%%%%%%%%%%%%%%%%%%%%%%%%%%%%%%%%%%%%%%%%%%%%%%%%%%

Discovering signals in large-scale modern data is like looking for needles in the haystack -- a geneticist locates genes that are associated with a disease \citep{efron2001empirical, leng2013ebseq}; 
% an astronomer ;
a neural scientist discovers differential brain activity 
\cite{perone2004false, schwartzman2008false};
a technology firm screens multiple potential innovations among thousands of A/B testings \citep{goldberg2017decision, azevedo2020b, guo2020empirical};
a personalized health system optimizes physical activities with a reinforcement learning algorithm by identifying immediate treatment effects \citep{liao2020personalized};
a manager picks the superstars and underdogs among sports teams \citep{brown2008season, jiang2010empirical, cai2019statistical};
% and in education \citep{abadie2019}; 
an economist estimates the effects of a large number of treatments \citep{heckman1984method, abadie2019choosing}.
In addition, these large-scale studies often collect data from multiple sources introducing heterogeneity among units.
Failing to account for both sparsity and heterogeneity will compromise both estimation and testing for the signals.
In this paper, we propose a novel Spike-and- Nonparametric mixture prior (SNP) – a spike component to promote sparsity and a flexible nonparametric component to capture the heterogeneous signals
and adopt the Bayesian machinery to provide sparse and heteroscedastic signals estimation and a multiple testing procedure.
% Through a simple proof w
We show our multiple testing procedure is able to control false discovery rate at desired levels and at the same time to achieve higher power. 
% In this paper, we propose a novel Spike-and-Nonparametric mixture prior (SNP) and an optimal multiple testing procedure for sparse and heteroscedastic signals estimation and testing -- a spike component to promote sparsity and a flexible nonparametric component to capture the heterogeneous signals. The multiple testing procedure is able to control false discovery rate at desired levels and at the same time to achieve higher power.

As a motivation for our model setting, suppose we have gene expression data where each gene $i= 1, 2, \ldots, n$ is measured among two groups, healthy subjects and cancer patients. 
The goal is to estimate the ``true'' difference in gene expression between two groups $\mu_i$  and to test which genes are ``truly'' differentially expressed.  
It is expected that only few genes express differentially and the standard deviation of each gene expression varies.
A naive approach to estimate $\mu_i$ is to take the difference of two groups' averages, $y_i$, which is the maximum likelihood estimator (MLE). To perform simultaneous testing, one can scale the mean difference $y_i$ by the pooled standard error $\sigma_i$ then apply a standard multiple testing procedure to $y_i/\sigma_i$, such as \cite{benjamini1995controlling, storey2002direct}, to identify significant genes.

It turns out that, unsurprisingly, the simple MLE solution is not the best. The MLE does not borrow information from each other and we should expect there exist commonalities among $\mu_i$. One proposal to borrow strength is to assume $\mu_i$ follows a common prior $\pi(\cdot)$. If we further assume $y_i \overset{\text{ind}}{\sim} p(y_i | \mu_i)$, then by the Bayes' rule, we obtain the posterior distribution $p(\mu_i|y_i) = p(y_i | \mu_i)\pi(\mu_i) / p(y_i)$. However, since the prior $\pi$  is unknown, we will estimate it by maximizing the marginal likelihood function. \cite{robbins1956sequential} introduced this method and named it the empirical Bayes method.
% and is dominated by the shrinkage estimator by \cite{james_stein}. 
The empirical Bayes method has a long-standing reputation of optimally borrowing information, and thus the crux of our problem is how to adaptively incorporate sparsity and heterogeneity into our model.

%%%%%%%%%%%%%%%%%%%%%%%%%%%%%%%%%%%%%%%%%%%%%%%%%%%%%%%%%%%%
%%%%%%%%%%%%%%%%%%%%%%%%%%%%%%%%%%%%%%%%%%%%%%%%%%%%%%%%%%%%
% \section{The Spike-and-nonparametric Mixture Prior}
%%%%%%%%%%%%%%%%%%%%%%%%%%%%%%%%%%%%%%%%%%%%%%%%%%%%%%%%%%%%
%%%%%%%%%%%%%%%%%%%%%%%%%%%%%%%%%%%%%%%%%%%%%%%%%%%%%%%%%%%%
\paragraph{The Spike-and-Nonparametric Mixture Prior (SNP)}
We now formulate the gene expression example into a canonical problem of estimating a high-dimensional mean vector from a single observation. Specifically, the observed vector $ \y = (y_1, y_2, \ldots, y_n)^\top$ satisfies 
\begin{equation}
\label{eq:y}
	y_i = \mu_i + \epsilon_i \quad \mbox{where}\quad  
	\epsilon_i \sim  N\left(0, \sigma^2_i \right),\, i = 1, \ldots, n
\end{equation}
where  $\bsigma=(\sigma_1,\ldots, \sigma_n)^\top$ are known and bounded. The goal is to estimate $\bmu = (\mu_1, \ldots, \mu_n)^\top$ with precision.

For this sparse and heteroscedastic setting, we introduce the family of Spike-and-Nonparametric mixture priors for $\bmu$,
\begin{equation}
	\label{eq:snp-fam}
	\pi \func{\mu_i | \omega, \gamma} =  \omega \psi(\mu_i | \lambda_0) + (1-\omega)\gamma(\mu_i)
\end{equation}
where the spike $\psi(x|\lambda_0)= \frac{\lambda_0}{2} e^{-\lambda_0 |x|}$ is a Laplace (double exponential) distribution assuming $\lambda_0$ is large, $\gamma(\cdot)$ is a smooth nonparametric probability density of non-zero $\mu_i$'s and $\omega \in [0,1]$ is the weight on the spike controlling the sparsity level.
% The mixture prior consists of a Laplace distribution at zero and some unknown distribution with density $\gamma(\cdot)$. 
When $\lambda_0\rightarrow \infty$, we obtained the dirac-mass nonparametric mixture prior DNP \eqref{eq:pi-mu} as a limiting special case of SNP \eqref{eq:snp-fam}, i.e.,
\begin{equation}
	\label{eq:pi-mu}
	\pi \func{\mu_i | \omega, \gamma} =  \omega \delta_0 \func{\mu_i} + (1-\omega) \gamma(\mu_i)
\end{equation}
where the spike is $\delta_0(\mu_i)$, a point mass of 1 at $\mu_i=0$. 
%Both SNP and DNP reflect our belief that some of $\mu_i = 0$.  BLAH BLAH BLAH
% The mixture parameter $\omega$ controls the level of sparsity and is considered as a hyper-parameter. 

Under this broad framework with the proposed spike-and-nonparametric mixture prior, we estimate the prior distribution of the location parameter $\bmu$ through an expectation-maximization (EM) algorithm \citep{dempster1977maximum, laird1978nonparametric}. The estimated prior distribution further enables us to derive the posterior distribution given the observations, which is a powerful tool for statistical estimation and inference. 

In sum, our proposed spike-and-nonparametric mixture prior has several merits.
\begin{enumerate}[leftmargin=*,itemsep=0mm]
\item \textbf{Shrinkage}. SNP inherits the shrinkage property of empirical Bayes methods for normal mean estimation. In addition, the shrinkage is multi-directional, towards zero for noises and towards their corresponding nearest means for signals;
%In addition, it is greedy due to the nonparametric portion and aims at the smallest risk. 
\item \textbf{Thresholding}. As a variant of the spike-and-slab priors, SNP is capable of adaptive thresholding to shrink some estimates to be exactly zero when using the posterior mode estimator;
\item \textbf{Sparsity Adaptive}. Both the shrinkage and thresholding properties of SNP adapt to different sparsity levels;
\item \textbf{Sparsity estimation}. SNP is able to estimate the sparsity well. The estimated sparsity can be then incorporated into  multiple testing procedures to increase the power;
\item \textbf{Multiple testing}. The adaptivity to the sparsity of SNP leads to a multiple testing procedure that controls false discovery rate (FDR) at desired levels while achieves higher power;
\item \textbf{Bayesian credible interval}. Based on the entire posterior of SNP, the Bayesian mechanism further enables us to provide credible intervals for uncertainty quantification.
% With the entire posterior of SNP, the Bayesian mechanism further provides credible interval for uncertainty quantification.
\end{enumerate}

Our work builds on a long tradition of empirical Bayes estimation and the spike-and-slab priors in Bayesian variable selection.
Empirical Bayes estimators adopt the Bayesian philosophy of prior while retaining the frequentist's optimal property of their counterparts, such as the James-Stein estimator and LASSO \citep{james_stein, tibshirani1996regression, park2008bayesian, carvalho2010horseshoe}. \cite{efron2014two} and \cite{efron2019bayes} summarized two modelling strategies -- the $g$-modelling that estimates the prior of $\mu$ and the $f$-modelling that estimates the marginal density of $y$. %%%%%%%%%%%%%%%
On the other hand, the family of spike-and-slab priors has been widely adopted in Bayesian variable selection. Although the structure of spike-and-slab priors is very similar to that of SNP \eqref{eq:snp-fam}, a mixture structure that consists of a spike and a slab, it is the function forms imposed on the spike and the slab that differentiate the priors, such as the parametric point-mass prior variants \citep{george_foster_biometrika_2000, johnstone_AS_2004, abramovich_AS_2006} and the continuous variants
% that consists two normal, student-$t$, and Laplacian mixtures 
\citep{george1993variable, ishwaran2003detecting, ishwaran2005spike, ishwaran2011consistency, rovckova2016spike, rovckova2018bayesian}. 	While the parametric mixture priors enjoy mathematical tractability and computational efficiency, the nonparametric variants show robust performance \citep{wand1994kernel, brown2009nonparametric, jiang2009general, raykar2010nonparametric, castillo2012needles}.
On the other hand, research in empirical Bayes credible interval is active 
\citep{van2017uncertainty, belitser2020empirical, rousseau2020asymptotic, castillo2020spike, banerjee2021bayesian}
but none is directly applicable to our prior.

As a simultaneous estimation strategy, it is natural to connect to the multiple testing theory \citep{benjamini1995controlling, benjamini2001control, storey2002direct, efron2004large, sun2007oracle, efron2012large}. The mixture structure of spike-and-slab priors has direct implications on the null and non-null groups. Under some structural assumptions of the spike and slab, theoretical guarantees of the corresponding empirical Bayes multiple testing have been established \citep{castillo2020spikebmt, abraham2021empirical}. Literature on simultaneous inference after selection is sparse \citep{yekutieli2012adjusted, yu2018adaptive, woody2018optimal}.

Most works focus on either estimation or testing, our approach does both in a unified framework. In addition, we provide uncertainty quantification for the estimation. Note that most of the methods mentioned above focus on homoscedastic error until recent extensions to the heteroscedastic case, both for empirical Bayes estimation \citep{xie2012sure, weinstein2018group, banerjeenonparametric, jiang2020general} and multiple testing \citep{lei2016adapt, ignatiadis2016data, fu2020heteroscedasticity}.
The closest state-of-the-arts include \cite{banerjeenonparametric} that focuses on estimation with a nonparametric prior using the $f$-modelling, which is unable to estimate sparsity or provide uncertainty quantification, 
\cite{jiang2020general} that adopts the $g$-modelling with a nonparametric prior using the convex approximation by \cite{koenker2013convex} and \cite{koenker2017rebayes} instead of the EM algorithm, 
the spike-and-slab LASSO \citep{rovckova2016spike} that imposes two Laplace distributions for the mixture prior with no nonparametric components,
and \cite{fu2020heteroscedasticity} that focuses on multiple testing adjusting for heteroscedasticity. Our method cross-fertilizes the merits of both Laplacian's ability to capture sparsity and the nonparametric empirical Bayes $g$-modeling's flexibility to model signals, and thus provides accurate sparsity estimation, signal estimation with uncertainty quantification, and an optimal multiple testing procedure. 

The rest of the paper is organized as follows. 
Section \ref{sec:setup} summarizes the Bayesian set up.
Section \ref{sec:est} describes an EM algorithm to estimate the unknown hyper-parameters in the prior. The estimated hyper-parameters are then plugged in to derive the posterior and to produce point estimates and credible intervals.
Section \ref{sec:multiple-testing} focuses on the multiple hypothesis testing problem and derives an optimal FDR control procedure based on the posterior. 
In Section \ref{sec:emp}, our empirical results of both simulations and a gene expression case study demonstrate that our procedure adapts to sparsity and heteroscedasticity better than its counterparts. 
We conclude the paper with discussions in Section \ref{sec:discussion}. 
% Some detailed derivations, proofs and empirical results are deferred to the Appendix.

%%%%%%%%%%%%%%%%%%%%%%%%%%%%%%%%%%%%%%%%%%%%%%%%%%%%%%%%%%%%
%%%%%%%%%%%%%%%%%%%%%%%%%%%%%%%%%%%%%%%%%%%%%%%%%%%%%%%%%%%%
\section{The Bayesian Setup}
\label{sec:setup}
%%%%%%%%%%%%%%%%%%%%%%%%%%%%%%%%%%%%%%%%%%%%%%%%%%%%%%%%%%%%
%%%%%%%%%%%%%%%%%%%%%%%%%%%%%%%%%%%%%%%%%%%%%%%%%%%%%%%%%%%%

%
%Given the model assumption \eqref{eq:y} and the DNP \eqref{eq:pi-mu} and SNP prior \eqref{eq:snp-fam},
%\vspace{-.15in}
\paragraph{Likelihood}
The likelihood of the parameters $\bmu$ 
given the independent observations $\y$ can be factorized into 
$L(\y|\bmu) = \prod_{i=1}^n p(y_i |\mu_i, \bsigma) = \prod_{i=1}^n \varphi(y_i |\mu_i, \sigma_i)$
where $\varphi$ is the density function for normal distribution. 
Note that the maximum likelihood estimator of $\bmu$ is the observation $\y$ itself.
The normality assumption can be relaxed to the entire exponential family. 
The case of correlated $\y$ can be addressed by a two-stage procedure as described in Section \ref{sec:discussion}.
\vspace{-.15in}
\paragraph{Prior} Our Spike-and-Nonparametric mixture prior (SNP) consists of   a Laplace distribution 
% $\psi(x|\lambda_0)= \frac{\lambda_0}{2} e^{-\lambda_0 |x|}$ 
as the spike and a nonparametric probability density function as specified in \eqref{eq:snp-fam}.
% $\gamma$, i.e., $$\pi \func{\mu_i | \omega, \gamma} =  \omega \psi(\mu_i | \lambda_0) + (1-\omega)\gamma(\mu_i)$$ as in \eqref{eq:snp-fam}. 
The Dirac-mass-and-Nonparametric mixture prior (DNP) replaces the Laplace distribution as a point mass at zero 
as specified in \eqref{eq:pi-mu}.
% i.e., $\pi \func{\mu_i | \omega, \gamma} =  \omega \delta_0 \func{\mu_i} + (1-\omega) \gamma(\mu_i)$ as in \eqref{eq:pi-mu}.
\vspace{-.15in}
\paragraph{Posterior} We assume that each $\mu_i$ follows independently from SNP. Given the hyper-parameter $\omega, \lambda_0$ and $\gamma$ of SNP, the posterior of $\bmu$ given the data $\y$ is
\begin{align}
p(\bmu | \y, \omega, \lambda_0, \gamma, \bsigma) &= \dfrac{\prod_{i=1}^n  \varphi(y_i | \bmu, \sigma_i)\pi(\bmu | \omega, \lambda_0, \gamma) }{m(\y | \omega, \lambda_0, \gamma, \bsigma)}, 
% \quad \mbox{ where}
\label{eq:posterior}
%\end{equation}
\end{align}
where the marginal density of the observed $\y$ is
\begin{align}
m(\y | \omega, \lambda_0, \gamma, \bsigma)  
% &= \prod_{i=1}^n m(y_i | \omega, \lambda_0, \gamma, \sigma_i) \\
&= \prod_{i=1}^n  \int \varphi(y_i |\mu_i, \sigma_i) \pi(\mu_i | \omega, \lambda_0, \gamma) d\mu_i. \label{eq:marginal} 
% &= \prod_{i=1}^n  \int \varphi(y_i | \mu_i, \sigma_i)
% \big[\omega \psi(\mu_i | \lambda_0) + (1-\omega) \gamma(\mu_i) \big]d\mu_i
% \nonumber 
\end{align}
The posterior \eqref{eq:posterior} can now be factorized into 
$p(\bmu | \y, \omega, \lambda_0, \gamma, \sigma) = \prod_{i=1}^n p(\mu_i | y_i, \omega, \lambda_0, \gamma, \sigma_i)$ where
\begin{align}
% \begin{split}
& \quad p(\mu_i | y_i, \omega, \lambda_0, \gamma, \sigma_i) \label{eq:posterior-i}     \\
&= \dfrac{ \omega \varphi(y_i | \mu, \sigma_i) \psi(\mu_i | \lambda_0) + (1-\omega)  \varphi(y_i | \mu, \sigma_i)\gamma(\mu_i)  }{m(y_i | \omega, \lambda_0, \gamma, \sigma_i)} \nonumber. 
% \end{split}
\end{align}

We now have all the Bayesian ingredients. Once we estimate all the 
unknown parameters in \eqref{eq:posterior-i}, we can 
produce point estimates as well as
credible intervals for  $\mu_i$'s from the posterior. 
We also obtain the posterior probability of
$\mu_i$ being zero which plays the key roles
in sparsity estimation and multiple testing.

%\paragraph{Multiple testing}
%The posterior probability of $\mu_i=0$ entails the likelihood of $\mu_i$ being in the non-signal group. 
%For SNP the posterior probability of $\mu_i=0$ is 
%\begin{equation}
%\label{eq:snp-post-0}
%p_i = \Prob(\mu_i \in [\delta_L, \delta_R] | y_i, \omega, \lambda_0, \gamma, \sigma_i) 
%%= \dfrac{p(y_i | \mu \in [\delta_L, \delta_R], \sigma_i) \cdot \hat{\pi}(\mu \in [\delta_L, \delta_R])}{m(y_i| \omega, \lambda_0, \gamma, \sigma_i)}	
%\end{equation}
% where $\delta_L$ and $\delta_R$ are the intersects of two mixtures, $\psi(\mu|\lambda_0)$ and $\gamma(\mu)$. 
%
%For DNP, the posterior probability of $\mu_i=0$ can written as
%\begin{equation}
%\label{eq:dnp-post-0}
%p_i = \Prob(\mu_i =0 | y_i,  \omega, \gamma, \sigma_i) 
%= \omega \varphi(y_i | 0, \sigma_i) / m(y_i | \omega, \gamma, \sigma_i)
%\end{equation} 
%where the marginal is
%$
%m(y_i | \omega, \gamma, \sigma_i) = 
%\omega \varphi(y_i | 0, \sigma_i) + (1-\omega) \int  \varphi(y_i | \mu_i, \sigma_i) \gamma(\mu_i) d\mu_i.
%$

%%%%%%%%%%%%%%%%%%%%%%%%%%%%%%%%%%%%%%%%%%%%%%%%%%%%%%%%%%%%
%%%%%%%%%%%%%%%%%%%%%%%%%%%%%%%%%%%%%%%%%%%%%%%%%%%%%%%%%%%%
\section{Sparsity Adaptive Empirical Bayes Estimation}
\label{sec:est}
%%%%%%%%%%%%%%%%%%%%%%%%%%%%%%%%%%%%%%%%%%%%%%%%%%%%%%%%%%%%
%%%%%%%%%%%%%%%%%%%%%%%%%%%%%%%%%%%%%%%%%%%%%%%%%%%%%%%%%%%%

In this section, we present our sparsity adaptive estimation.
We estimate the unknown high-dimensional hyper-parameters in the prior by maximizing the marginal likelihood \eqref{eq:marginal}.
The posterior of $\bmu$ is then obtained by plugging in the estimated prior.
With posterior of $\bmu$, we provide the sparsity estimation and use posterior mean or posterior mode as a point estimate of $\bmu$. 
Since we estimate the entire posterior of $\bmu$, the highest posterior density (HPD) or equal-tailed credible intervals for $\bmu$ can be readily constructed. 

\subsection{Prior estimation using Expectation-Maximization Algorithm}
%%%%%%%%%%%%%%%%%%%%%%%%%%%%%%%%%%%%%%%%%%%%%%%%%%%%%%%%%%%%

In order to facilitate our EM Algorithm for the nonparametric prior estimation, we discretize the nonparametric function $\gamma$ into M many equal-length mesh with grid points, $\min(\y)-2 \cdot sd(\y) < \tau_j < \tau_{j+1} < \ldots <\tau_{M} < \max(\y) + 2 \cdot sd(\y)$, and denote $\pi_j$ as the weights of the support points $\tau_j$ for $j =1 ,\ldots, M$. 
% $\pi_j= \Prob\left( \mu \in [\tau_j, \tau_{j+1})\right)$ for $j =1 ,\ldots, M$. 
We impose zero as one of the grid points and $\sum_j \pi_j  =1$.
% We use a histogram-type estimator with bin weight on the left point.
%%%%%%%%%%%%%%%%%%%%%%%%%%%%%%%%%%%%%%%%%%%%%%%%%%%%%%%%%%%%
\paragraph{Dirac Prior}
%%%%%%%%%%%%%%%%%%%%%%%%%%%%%%%%%%%%%%%%%%%%%%%%%%%%%%%%%%%%
% With a flexible prior (\ref{prior-mulnomial}), it is important to estimate the prior $\bm \pi$ well and efficiently. 
After the discretization of $\gamma$,
% With the nonparametric function $\gamma$ discretized,
we use an EM algorithm to obtain the maximum likelihood estimate of the Dirac-Nonparametric Prior (DNP). Note the under discretization, DNP is equivalent to
%\begin{equation}
%	\label{prior-mulnomial}
$
	\pi \func{\mu_i | \bm \pi} \stackrel {i.i.d} {\sim} \text{Mul} (\bm \pi)
	$
%\end{equation} 
where $\bm \pi = \{\pi_j\}_{j=1}^M$ are the hyper-parameters and $\pi(0)=w$ is the sparsity. 
The maximum likelihood estimator maximizes the following conditional log-likelihood function of $\y$ given $\bm \pi$ 
 \begin{equation}
 	\label{marginal-lik}
	     	\ell(\y | \bm\pi) = \sum_{i=1}^n \log\Bigg\{ \sum_{j=1}^M  p(y_i | \mu_j, \sigma ^2_i) ~ \pi_j \Bigg\}.
 \end{equation}
% 
% The maximization is a high dimensional problem since M is set to be rather large. We use EM algorithm proposed by \cite{jiang2009general} and adopt earlier stopping rules. The major differences between \cite{jiang2009general} and this project are that they do not handle heterosedastic set up and we have an ealier stopping rule in the EM algorithm to avoid overfitting. \cite{castillo2012needles} showed the posterior concentration properties of such prior but only in the homoscedastic models. 
 
 \begin{theorem}
 \label{thm:dnp-em}
Denote estimate of $\pi$ in round $t$ as $\piold$ and define 
% 	\begin{equation}
% 		\label{cond-z}
 		$\piold_{j,i} = \frac{p(y_i| \tau_j)~ \piold_j}{\sum_{m=1}^M p(y_i | \tau_m)~ \piold_m}$.
%	\end{equation}
The EM algorithm updates $\bm\pi=(\pi_1, \ldots, \pi_M)$ by
%\begin{equation}
%\label{em-sol}
$
	\pinew_j  = \frac{1}{n}\sum_{i=1}^{n}\piold_{j,i}. 
	$
%\end{equation}
If $M=M_n\rightarrow \infty$ , the limit of the algorithm is consistent in the sense of weak convergence.

%The above EM estimators are in its most basic for5m, consistent (Kiefer and Wolfowitz \cite{kiefer1956consistency}).  That is, denoting the NPMLE estimate of the true prior $\pi$ by $\hat{\pi}_{MLE}$, we have under relatively mild regularity and identifiability conditions that
%$$  \hat{\pi}_{MLE} \stackrel{n \rightarrow \infty}{\longrightarrow} \pi.   $$
 \end{theorem}

%%%%%%%%%%%%%%%%%%%%%%%%%%%%%%%%%%%%%%%%%%%%%%%%%%%%%%%%%%%%
\paragraph{Laplacian Spike Prior}
%%%%%%%%%%%%%%%%%%%%%%%%%%%%%%%%%%%%%%%%%%%%%%%%%%%%%%%%%%%%

As a continuum between the Dirac Mass-Nonparametric prior and Laplacian prior,
SNP retains the mixture structure that consists of a Laplacian spike centered at zero, replacing the point mass, and a nonparametric portion. The maximum likelihood estimator is equivalent to solving the following optimization problem, i.e.,
\begin{equation}
\label{eq:snp}
\argmax_{\omega, \lambda_0, \bm\gamma} \sum_{i=1}^n \log \int \varphi(y_i - \mu) \pi(\mu | \omega, \lambda_0, \gamma) d\mu 
\end{equation}
where the SNP prior $\pi(\mu | \omega, \lambda_0, \gamma) = \omega \psi(\mu | \lambda_0) + (1-\omega)\gamma(\mu)$. 
%$\omega \in [0,1]$ is the weight on the spike, $\psi(x|\lambda_0)= \frac{\lambda_0}{2} e^{-\lambda_0 |x|}$ is a Laplace (double exponential) distribution, and $\gamma(\cdot)$ is the probability density of non-zero $\mu_i$'s. The SNP becomes a mixture of a Laplace distribution at 0 and some unknown distribution with density $\gamma(\cdot)$. We estimate the unknown prior function $\pi$ using the empirical Bayes method. 
% The GMLEB becomes a linear combination of
%the parametric and the discretized nonparametric portion, i.e.
%\[\begin{split}
%\argmax_{\omega, \lambda_0, \bm\gamma} & 
%\prod_{i=1}^n 
%\br*{\omega \int \varphi(y_i-u) \psi(u | \lambda_0) du + 
%(1-\omega) \sum_{j=1}^M \varphi(y_i-\tau_j) \gamma_j}	\\
%s.t. & \quad \sum_j \gamma_j = 1 
%\mbox{ and } 
%\gamma_j \geq 0 \\
%& \quad 0 \leq \omega \leq 1 .
%\end{split}\]
For the ease of computation, we discretize the prior $\pi(\mu | \omega, \lambda_0, \gamma)$ on the grid points. The optimization problem \eqref{eq:snp} becomes 
\begin{align}
% \begin{split}
\argmax_{\omega, \lambda_0, \bm\gamma} & 
\sum_{i=1}^n 
\log \Bigg\{\sum_{j=1}^M 
\varphi(y_i-\tau_j)
\big[\omega \psi(\tau_j | \lambda_0) + (1-\omega) \pi_j\big]\Bigg\}	 \nonumber \\
s.t & \quad   \sum_j \pi_j = 1, \quad  \pi_j \geq 0, \quad \lambda_0 >0 .\label{eq:snp-discrete}
% \end{split}	
\end{align}
To facilitate the EM algorithm, we introduce latent variable 
\begin{align}
  Z | \bm\theta \sim Multi(M, \bm\theta)   
\end{align}
% Z | \bm\theta \sim Multi(M, \bm\theta) 
% ~~~ \mbox{where } \Prob(Z = j) = \theta_j = \omega \psi (\tau_j) + (1-\omega)\pi_j \mbox{ and } \sum_j \theta_j = 1.
where $\theta_j = \Prob(Z = j) =  \omega \psi (\tau_j) + (1-\omega)\pi_j \mbox{ and } \sum_j \theta_j = 1$.
%Denote $\bm\pi=(\pi_1, \ldots, \pi_M)$.
We maximize the discretized maximum likelihood \eqref{eq:snp-discrete} indirectly, by treating the latent variables $\bm Z$ as ``missing data'' and maximizing the ``complete-data'' log-posterior $\log \pi(\omega, \lambda_0, \bm\pi, \Z | \y)$. Following the EM recipe, the E-step replaces ``complete-data'' log-posterior by its conditional expectation with respect to $\bm Z$ given the observed data $\y$ and the estimate $(\omega^{(t)}, \lambda_0^{(t)}, \bm\pi^{(t)})$ at the $t$ step, and then the M-step maximizes the conditional expected complete-data log-posterior with respect to $(\omega, \lambda_0, \bm\pi)$.
 
To be specific, in the E-step, the conditional expectation with respect to $\bm Z$ given the observed data $\y$ and the estimates $(\omega^{(t)}, \lambda_0^{(t)}, \bm\pi^{(t)})$ is
\begin{align}
% \begin{split}
& \E_{\bm Z | \omega^{(t)}, \lambda_0^{(t)}, \bm\pi^{(t)}, \bm y}
\bk*{\log \pi(\omega, \lambda_0, \bm\pi | \omega^{(t)}, \lambda_0^{(t)}, \bm\pi^{(t)}, \y)} \nonumber \\
&= \sum_{i=1}^n \sum_{j=1}^M \frac{\varphi(y_i - \tau_j) \theta_j^{(t)} }{\sum_k \varphi(y_i - \tau_k) \theta_k^{(t)}}
 \bigg[\log\big(\varphi(y_i - \tau_j)\big) +  \nonumber \\
 & \hspace{1in} \log\big(\omega \psi(\tau_j | \lambda_0) + (1-\omega) \pi_j\big)\bigg] \label{em:e-step}
% \end{split}	
\end{align}

In the M-step, we maximize the conditional expectation with  respect to $(\omega, \lambda_0, \bm\pi)$, i.e.,
\begin{align}
% \begin{split}
&	\argmax_{\omega, \bm\pi, \lambda_0} 
	\E_{\bm Z | \omega^{(t)}, \lambda_0^{(t)}, \bm\pi^{(t)}, \bm y}
{\log \pi(\omega, \lambda_0, \bm\pi | \omega^{(t)}, \lambda_0^{(t)}, \bm\pi^{(t)}, \y)} \nonumber \\
%	- \sum_{i=1}^n \sum_{j=1}^M  \frac{\varphi(y_i - \tau_j) \theta_j^{(t)} }{\sum_k \varphi(y_i - \tau_k) \theta_k^{(t)}}
%  \bigg[\log\big(\varphi(y_i - \tau_j)\big) + \log\big(\omega \psi(\tau_j | \lambda_0) + (1-\omega) \pi_j\big)\bigg]  \\
  & \mbox{ s.t. }  \sum_j \pi_j = 1, \pi_j \geq 0, \lambda_0 > 0. \label{opt:max}
% \end{split}
\end{align}
The optimization problem can be solved by setting the partial derivatives of the Lagrangian of $\E_{\bm Z | \omega^{(t)}, \lambda_0^{(t)}, \bm\pi^{(t)}, \bm y}
\bk*{\log \pi(\omega, \lambda_0, \bm\pi | \omega^{(t)}, \lambda_0^{(t)}, \bm\pi^{(t)}, \y)}$ to each parameter to zero. Denote $\pi_{\cdot,j}^{(t)} = \Prob\p*{Z_i = j | \omega^{(t)}, \lambda_0^{(t)}, \bm\pi^{(t)}, \bm y} = \sum_i \frac{\varphi(y_i - \tau_j) \theta_j^{(t)} }{\sum_k \varphi(y_i - \tau_k) \theta_k^{(t)}}$. We obtain the analytical update step for $\pi_j$ as
\begin{align}
\pi_j 
&= \max\left(0, \frac{\pi_{\cdot, j}^{(t)}}{(1-\omega) \sum_j \pi_{\cdot, j}^{(t)}}
- \frac{\omega}{1-\omega} \psi(\tau_j | \lambda_0)  \right)
% ~~~\mbox{for } j=1,\ldots, M 
\end{align}
and normalize $\sum_j \pi_j = 1$.
For $\lambda_0$ and $\omega$, find the roots of the following equations as the solution
\begin{align}
& \sum_{j=1}^M \pi_{\cdot,j}^{(t)}\frac{\frac{\omega}{2} \p*{1 -\lambda_0 |\tau_j|} e^{-\lambda_0 |\tau_j|} }{\omega \psi(\tau_j | \lambda_0) + (1-\omega) \pi_j} = 0 \\
& \sum_{j=1}^M \pi_{\cdot,j}^{(t)}\frac{\psi(\tau_j|\lambda_0) - \pi_j}{\omega \psi(\tau_j | \lambda_0) + (1-\omega) \pi_j} = 0.
\end{align}
%Details can be found in the Appendix \ref{app:snp}.

%%%%%%%%%%%%%%%%%%%%%%%%%%%%%%%%%%%%%%%%%%%%%%%%%%%%%%%%%%%%
\subsection{Posterior Distribution, Estimation and Inference}
%%%%%%%%%%%%%%%%%%%%%%%%%%%%%%%%%%%%%%%%%%%%%%%%%%%%%%%%%%%%

For notational simplicity, we denote the estimated prior for both DNP and SNP as $\hat\pi$.
We obtain the posterior distribution by plugging in the estimated prior $\hat\pi$, i.e.,
\begin{equation}
	\label{eq:pos-mul}
\hat{p}(\mu | y_i, \textcolor{blue}{\hat{\pi}}, \sigma_i) = \dfrac{\varphi(y_i | \mu, \sigma_i) \textcolor{blue}{\hat{\pi}(\mu)}}{\hat m(y_i | \sigma_i)}.
\end{equation}
With the posterior distribution, we provide solutions to point estimators with credible intervals and sparsity estimation as follows.

%%%%%%%%%%%%%%%%%%%%%%%%%%%%%%%%%%%%%%%%%%%%%%%%%%%%%%%%%%%%
\subsubsection{Point Estimation}
%%%%%%%%%%%%%%%%%%%%%%%%%%%%%%%%%%%%%%%%%%%%%%%%%%%%%%%%%%%%
%The foremost problem is to estimate the location parameter $\{\mu_i\}_{i=1}^n$ based on the observations $\{y_i\}_{i=1}^n$. Consider the squared error loss $L(\bmu,\delta(\x))=\|\delta(\x)-\bmu\|^2$. The corresponding risk function will be $R(\bmu,\delta(\x))=E_{\mu}L(\bmu,\delta(\x))$. By definition, the Bayes estimator $\delta_B$ minimized the posterior risk. More specifically, $\delta_B=\argmin_{d\in\mathcal{R}^n}E_{\mu}[L(\bmu,d)|\x]$. Take the derivative of $E_{\mu}[L(\bmu,d)|\x]$ with respect to $d$ and set it equal 0. This leads to $\delta_B=E[\bmu|\x]$, which motivates us to consider the posterior mean. Since the prior distribution of $\bmu$ is now formulated through $\bpi$, we will propose the following estimator in the current paper. 

%We estimate the normal mean by posterior mean or posterior mode. Posterior mode has the hard-thresholding effect that shrinks some estimates to 0.

\emph{Posterior Mean.}
The posterior mean estimator is defined as $ \hat{\mu}_i  = \sum_{j=1}^{M} \tau_j \; \hat{p}(\mu = \tau_j | y_i, \hat{\pi}, \sigma_i)$. 
%We use the posterior mean as our point estimates for $\mu_i$'s
%\begin{equation}
%	\label{pos-mean}
% \hat{\mu}_i  = \sum_{j=1}^{M} \tau_j \; \hat{p}(\mu = \tau_j | y_i, \hat{\pi}, \sigma_i^2). 
% \end{equation}
Our posterior mean estimator has a multi-directional shrinkage property, towards zero for noises induced by the spike and towards their corresponding nearest centers for signals induced by the flexible nonparametric component. 
% Illustrated in Figure \ref{fig:multi-shrink} as an example, the posterior mean estimate $\hat{\mu}$ ($\red{\times}$) shrinks the observed $y$ ($\bigcirc$) towards its corresponding mean $\mu$ ($\blue{\bullet}$), which is -5, 0 or 5 respectively.

% \begin{wrapfigure}{r}{0.45\textwidth}
%     \includegraphics[width=0.4\textwidth]{dir/multi-shrinkage}
%     \vspace{-.2in}
%   \caption{\footnotesize{The multi-directional shrinkage property of the posterior mean estimator.}}
%   \label{fig:multi-shrink}
%   \vspace{-.5in}
% \end{wrapfigure}

\emph{Posterior Mode.} The posterior mode estimator is defined as the $\mu$ corresponding to the global mode of the posterior.
%$ \hat{\mu}_i = (|y_i| - \lambda^*( \hat{\mu}_i))_+ \text{sign}(y_i)$
%where $\lambda^*( \hat{\mu}_i) = -\lambda_0  p(\mu_i)  + (1-p(\mu_i) ) + 
%\frac{(1-\omega)(\gamma' - \gamma)}{\omega \psi(\mu_i | \lambda_0) + (1-\omega)\gamma(\mu_i)}$ and $p(\mu_i) = \frac{\omega \psi(\mu_i | \lambda_0)}{\omega \psi(\mu_i | \lambda_0) + (1-\omega)\gamma(\mu_i)}$.
The posterior mode estimator has the soft-thresholding effect that shrinks some estimates to 0 due to the spike around zero. Meanwhile, the nonparametric specification also guarantees more flexibility in capturing the data underlying structure. The above two components help us achieve a better mean squared error performance, as illustrated in the numerical studies. 

%%%%%%%%%%%%%%%%%%%%%%%%%%%%%%%%%%%%%%%%%%%%%%%%%%%%%%%%%%%%
\subsubsection{Sparsity Estimation}
%%%%%%%%%%%%%%%%%%%%%%%%%%%%%%%%%%%%%%%%%%%%%%%%%%%%%%%%%%%%
In the current paper, we use $\omega$ to denote the proportion of zero $\mu_i$'s, through the weight in the spike-and-nonparametric prior. This sparsity $\omega$ is an important quantity in the statistical inference. For example, in the well-known multiple testing procedures such as \cite{benjamini1995controlling} and \cite{storey2002direct}, the sparsity information is used in constructing the procedures. In practice, such sparsity information is usually unknown. A more accurate estimate of the sparsity will improve the power of the multiple testing procedure, as demonstrated by \cite{benjamini2006adaptive}. We will discuss such applications in multiple testing with more details in Section \ref{sec:multiple-testing}. Focusing on the sparsity estimation itself, we propose the following sparsity estimator for DNP and SNP.

With DNP, the posterior probability of $\mu_i=0$ follows immediately from the posterior distribution:
\begin{equation}
\label{eq:dnp-post-0-hat}
	 \hat{p}(\mu_i = 0 | y_i, \hat{\pi}, \sigma_i^2) = \dfrac{p(y_i | \mu = 0, \sigma_i^2) \cdot \hat{\pi}(\mu = 0)}{\hat{p}(y_i | \hat{\pi}, \sigma_i^2)}.
\end{equation}
We can then estimate the sparsity by the mean of 
%\begin{equation}
%\label{eq:dnp-w-hat}
$	\hat\omega = \frac{1}{n} \sum_{i=1}^n \hat{p}(\mu_i = 0 | y_i, \hat{\pi}, \sigma_i^2)$.
%\end{equation}

For SNP, the posterior probability of $\mu_i=0$ is 
\begin{align}
\label{eq:snp-post-0-hat}
	&\hat{p}(\mu_i \in [\delta_L, \delta_R] | y_i, \hat{\pi}, \sigma_i^2) = \nonumber \\
	& \dfrac{p(y_i | \mu \in [\delta_L, \delta_R], \sigma_i^2) \cdot \hat{\pi}(\mu \in [\delta_L, \delta_R])}{\hat{p}(y_i | \hat{\pi}, \sigma_i^2)}
\end{align}
where $\delta_L$ and $\delta_R$ are the intersects of two mixtures, $\psi(\mu|\hat\lambda_0)$ and $\hat\gamma(\mu)$, that are closest to 0 from the left and the right respectively.
    Specifically,
    $\delta_L$ is the closest grid point to the largest root of $\gamma(\mu) = \psi(\mu|\lambda_0)$ among $\mu < 0$. If there exists no root that is less than 0, we take $\delta_L = -\infty$.  
    Similarly, $\delta_R$ is the closest grid point to the smallest root of $\gamma(\mu) = \psi(\mu|\lambda_0)$ among $\mu > 0$. If there exists no root that is larger than 0, we take $\delta_R=\infty$.
Similarly, the sparsity is estimated as
%\begin{equation}
%\label{eq:snp-w-hat}
$\hat\omega = \frac{1}{n} \sum_{i=1}^n 
\hat{p}(\mu_i \in [\delta_L, \delta_R] | y_i, \hat{\pi}, \sigma_i^2).$
%\end{equation}

%%%%%%%%%%%%%%%%%%%%%%%%%%%%%%%%%%%%%%%%%%%%%%%%%%%%%%%%%%%%
\subsubsection{Credible Interval}
%%%%%%%%%%%%%%%%%%%%%%%%%%%%%%%%%%%%%%%%%%%%%%%%%%%%%%%%%%%%
We construct the the credible interval for $\mu_i$ by 
choosing the posterior $1-\alpha$ equal-tailed interval,
\begin{equation}
	\label{pos-int}
 \hat{p}(\mu_i \in [l,u] | y_i, \hat{\pi}, \sigma_i^2) = \dfrac{p(y_i | \mu \in [l,u], \sigma_i^2) \hat{\pi}(\mu \in [l,u])}{\hat{p}(y_i | \hat{\pi}, \sigma_i^2)}
 \end{equation}
 where $l$ and $u$ denote the lower and upper quantiles, i.e. the $\alpha/2$-th and $(1-\alpha/2)$-th quantiles of the posterior.
 
% Most empirical Bayes $f$-modeling strategies cannot provide any uncertainty quantification by nature unless imposing parametric forms. 
% This again shows the advantage of our prior estimation.

%%%%%%%%%%%%%%%%%%%%%%%%%%%%%%%%%%%%%%%%%%%%%%%%%%%%%%%%%%%%
%%%%%%%%%%%%%%%%%%%%%%%%%%%%%%%%%%%%%%%%%%%%%%%%%%%%%%%%%%%%
\section{Multiple Testing}
\label{sec:multiple-testing}
%%%%%%%%%%%%%%%%%%%%%%%%%%%%%%%%%%%%%%%%%%%%%%%%%%%%%%%%%%%%
%%%%%%%%%%%%%%%%%%%%%%%%%%%%%%%%%%%%%%%%%%%%%%%%%%%%%%%%%%%%

%After estimation, it is natural to test which signals are significant. 
%\cite{efron2012large} shows that large-scale simultaneous testing often introduces \textit{selection bias}. 
In a multiple testing framework, we want to simultaneously test
\begin{equation}
H_{i0}: \mu_i=0 \quad\quad\text{vs.}\quad\quad H_{ia}: \mu_i\neq0
\end{equation}
where $i=1,\cdots,m$ for the $m$ hypotheses. In our setting, $m=n$ where $n$ is the number of observations. 
%When we assume sparsity for the normal mean vector, that is, a proportion of the unknown means are zero, such multiple testing will be powerful for detecting the significance in the location parameter. 

False Discovery Rate (FDR) was introduced by \cite{benjamini1995controlling} and defined as the expected proportion of falsely rejected null hypotheses among all of the rejected.
The classification of tested hypotheses can be summarized in Table \ref{tbl:ar}. Correspondingly, $FDR=E[\frac{V}{R}]$ is usually the main focus in the multiple testing problem, where we use $0/0=0$ for convenience.

% \begin{wraptable}{r}{7cm}
% \vspace{-0in}
% \caption{Classification of tested hypotheses}\label{tbl:ar}
\begin{table}[h!!]
\begin{center}\caption{Classification of tested hypotheses}\label{tbl:ar}
  \begin{tabular}{cccc}
         \hline\hline
                       &Number    &Number    &\\
         Number of     &not rejected  &rejected  &Total\\
         \hline
         True Null     &$U$         &$V$      &$m_0$\\
         False Null    &$T$         &$S$      &$m_1$\\
         \hline
                       &$m-R$       &$R$      &$m$\\
         \hline
   \end{tabular}
\end{center}
\end{table}
% \vspace{-.1in}
% \end{wraptable} 
There are two major testing procedures for controlling FDR based on $p$-values.
 One is to compare the $p$-values with a data-driven threshold as in \cite{benjamini1995controlling}.  Specifically, let
$p_{(1)} \leq p_{(2)} \leq \cdots \leq p_{(p)}$ be the ordered observed $P$-values of $m$ hypotheses. Define $k = \text{max}\{i: p_{(i)} \leq i\alpha/m \}$ and reject $H_{(1)}^0, \cdots, H_{(k)}^0$, where $\alpha$ is a specified control rate. 
The other related approach is to find a threshold $t$ so that the estimated FDR is no larger than $\alpha$ \citep{storey2002direct}.  To find a common threshold, let $\mathrm{\widehat{FDR}}(t) = \widehat{m}_0t/(R(t)\vee 1)$, where $R(t)$ is the number of total discoveries with the threshold $t$ and $\widehat{m}_0$ is an estimate of $m_0$. \cite{storey2002direct} first estimated $\widehat{m}_0=(m-r(\lambda))/(1-\lambda)$ where $r(\lambda)=\#\{p_{(i)}\leq\lambda\}$, then solve $t$ such that $\mathrm{\widehat{FDR}}(t)\leq\alpha$. The parameter $\lambda$ can be selected by cross validation.
%Storey's procedure has been proved to asymptotically control the FDR when the test statistics are weakly dependent. The equivalence between the two methods has been theoretically studied by Storey, Taylor \& Siegmund (2004) and Ferreira \& Zwinderman (2006).

A related yet different strategy is to measure the likelihood of each $\mu_i=0$ conditional on the test statistics -- the smaller the likelihood, the more confident to reject the hypothesis. Such measure of significance is connected to optimality from a decision-theoretic perspective. Specifically, an optimal testing procedure can be constructed to minimize the objective function $\E(T)$, subject to $R=k$ for a positive integer number $k$. That is, given the number of total discoveries, we want to minimize the averaged number of false non-discoveries. 

Suppose we consider a decision vector $\a=(a_1,\cdots,a_m)$ where $a_i=1$ if we reject the $i$-th hypothesis and $a_i=0$ otherwise. A false discovery can be expressed as $a_i \I_{\mu_i= 0}$ where $\I$ is an indicator function, and similarly a false non-discovery as $(1-a_i)\I_{\mu_i\neq0}$. Correspondingly, the objective function can be written as
\begin{equation}\label{opt}
\min_{(a_1,\cdots,a_m)}\E\Big[\sum_{i=1}^m(1-a_i)\I_{\mu_i\neq0}\Big]
\quad \text{s.t.} \quad \sum_{i=1}^ma_i=k.
\end{equation}

\begin{proposition}\label{prop1}
Setting $a_i=1$ for the smallest $k$ of $p(\mu_i=0|y_i)$ obtains the optimal solution to (\ref{opt}). 
\end{proposition}
%
%\noindent{\bf Proof of Proposition \ref{prop1}}: The expectation in (\ref{opt}) is over the distribution of $\Y$ and the prior distribution of $\bmu$. It is equivalent to minimize the expectation of loss function conditional on $\Y$, that is, minimizing
%\begin{eqnarray}\label{eq20}
%&&\min_{(a_1,\cdots,a_m)} L(\a) = \sum_{i=1}^m\Big[(1-a_i)P(\mu_i\neq0|\Y)\Big]\\
%\text{subject to} && \sum_{i=1}^ma_i=k. \nonumber 
%\end{eqnarray}
%
%After some algebra, (\ref{eq20}) can be re-arranged as 
%\begin{equation*}
%L(\a)=\sum_{i=1}^m(1-P(\mu_i=0|\Y))+\sum_{i=1}^ma_i[P(\mu_i=0|\Y)-1]. 
%\end{equation*}
%Note that $L(\a)$ is increasing when the second term is increasing. Therefore, to minimize $L(\a)$ subject to $\sum_{i=1}^ma_i=k$, for the smallest $k$ values of $P(\mu_i=0|\Z)$, we set the corresponding $a_i=1$, and the rest of the $a_i$'s are set as 0. The proof is now complete. 

This motivates us to consider the following multiple testing procedure: sort $p_i$, the posterior probability of $\mu_i$ being null in a nondecreasing order, and denote the sorted $p_i$'s as $p_{(1)}, \cdots, p_{(p)}$. Then a conditional FDR given $\y$ can be expressed as
\begin{equation}\label{conFDR}
\frac{E[\sum_{i=1}^ma_i\I_{\mu_i=0}|\y]}{\sum_{i=1}^ma_i}=\frac{\sum_{i=1}^ma_ip_i}{\sum_{i=1}^ma_i}=\frac{1}{k}\sum_{i=1}^kp_{(i)}. 
\end{equation}
Choose the largest $k$ such that $k^{-1}\sum_{i=1}^kp_{(i)}\leq\alpha$ for a pre-determined $\alpha$ level. If the conditional FDR is controlled at $\alpha$ level, the FDR is also controlled at $\alpha$ level. It is worth mentioning that the conditional probability $\Prob(\mu_i=0|y_i)$ is the local fdr as in \cite{efron2007size}, and the form in (\ref{conFDR}) has also been considered in the literature with different motivations, e.g., \cite{sun2007oracle}, \cite{tangzhang2007}, \cite{sarkar2008general}, etc. 
%Since we assume the mixture model for the parameter $\mu_i$, because of the independence of $y_i$'s, $p_i$ can be further expressed as 
%\begin{equation*}
%p_i=\frac{\pi_0 \phi(y_i)}{\pi_0\phi(y_i)+(1-\pi_0)g(y_i)}. 
%\end{equation*}
Combining with our nonparametric empirical Bayes estimate of the density function, we propose the following multiple testing procedure. 

\paragraph{FDR control: NEB-OPT} The non-parametric prior empirical Bayes rule for significance level $\alpha$.

%1. For each observation $y_i$, compute the estimate $\hat{p}_{i}(y_i)$ for the posterior probability $\Prob(\mu_i=0 | y_i)$ using the formula
%\begin{equation}
%\hat{p_i}_{NB}(y_i)=\frac{\hat{\pi_0} \phi(y_i)}{\hat{\pi_0}\phi(y_i)+(1-\hat{\pi_0})\hat{g}(y_i)},
%\end{equation}
%where $\hat{\pi_0}$ stands for the estimate of $\pi_0=Prob(\mu_i=0)$, $\hat{g}(y_i)$ is an estimate of the marginal density of $y_i$ given that $\beta_i \neq 0$, and $\phi(y_i)$ is the value of the $N(0,1)$ density at $y_i$.

%2. Divide the interval between $z^{min}$ and $z^{max}$ into $N-1$ subintervals of equal length of the form \{$[x_1, x_2]$, $[x_2, x_3]$, ... , $[x_{N-1}, x_N]$\}, with $x_1=z^{min}$ and $x_N=z^{max}$.  Use the estimates computed in Step 1 above to find estimates $\hat{p}_{NB}(x_k)$ as follows:
%\begin{equation}
%\hat{p}_{NB}(x_1)=\hat{p}_{NB}(z^{min}),
%\end{equation}
%\begin{equation}
%\hat{p}_{NB}(x_N)=\hat{p}_{NB}(z^{max}),
%\end{equation}
%and, for $k \in \{2, ..., N-1\}$,
%\begin{equation}
%\hat {p}_{NB}(x_k) = \hat{p}_{NB}(z^{k, left})+\frac{\hat{p}_{NB}(z^{k, right})-\hat{p}_{NB}(z^{k, left}) }{z^{k, right}-z^{k,left}} (x_k - z^{k,left}), 
%\end{equation}
%where $z^{k, left}=max \{z_i \, s.t. \, z_i \leq x_k \}$ and $z^{k,right}=min\{z_i \, s.t. \, z_i \geq x_k\}$. \\
1. For each  $y_i$, estimate $\hat{p}_{i}(y_i)$ using the posterior probability of $\mu_i = 0$ of SNP \eqref{eq:snp-post-0-hat} or DNP \eqref{eq:dnp-post-0-hat}.

2.  Order the values $\hat{p}_{i}(y_i)$ computed in Step 1 from smallest to largest and denote the ordered list as $\{\hat {p}_{(1)}, ..., \hat{p}_{(m)} \}$. Let
%\begin{equation}
$K=\max \{k: \sum_{j=1}^{k} \hat{p}_{(k)}/k \leq \alpha \}.$
%\end{equation}
Reject the corresponding $K$ hypotheses. 

In addition to our procedure, our sparsity estimation $\hat\omega$ is able to improve BH and Storey's procedure as a better estimate of the proportion of zeros.
For BH procedure, the FDR is bounded by $\frac{m_0}{m}\alpha$ theoretically. BH procedure can be very conservative when $m_0$ is much smaller than $m$. We can incorporate the sparsity estimation $\hat\omega$ by adapting the rejection level as follows: Define $k = \text{max}\{i: p_{(i)} \leq i\alpha/(m\widehat{w}) \}$ and reject $H_{(1)}^0, \cdots, H_{(k)}^0$. 
For Storey's procedure, the estimate of the proportion of zeroes is based on a tuning parameter $\lambda$ and thus the performance of the testing procedure is sensitive to the choice of $\lambda$. 
We can adapt Storey's procedure by our estimate of the sparsity. Particularly, let $\mathrm{\widehat{FDR}}(t) = \widehat{w}mt/(R(t)\vee 1)$, then solve $t$ such that $\mathrm{\widehat{FDR}}(t)\leq\alpha$.

%%%%%%%%%%%%%%%%%%%%%%%%%%%%%%%%%%%%%%%%%%%%%%%%%%%%%%%%%%%%
%%%%%%%%%%%%%%%%%%%%%%%%%%%%%%%%%%%%%%%%%%%%%%%%%%%%%%%%%%%%
\section{Empirical Results}
\label{sec:emp}
%%%%%%%%%%%%%%%%%%%%%%%%%%%%%%%%%%%%%%%%%%%%%%%%%%%%%%%%%%%%
%%%%%%%%%%%%%%%%%%%%%%%%%%%%%%%%%%%%%%%%%%%%%%%%%%%%%%%%%%%%

In this section, we demonstrate the adaptivity of DNP and SNP to different levels of heteroscedasticity, signal strength and sparsity by simulation and apply our estimation and testing procedures to a gene expression case study.
More detailed results can be found in the Appendix.

\paragraph{Simulated data}
% The goal of the following design is to investigate the adaptivity of DNP and SNP to different levels of heteroscedasticity, in addition to signal strength and sparsity level. 
%Therefore, we vary not only the signal strength $V$ and the sparsity level $w$ as in the homoscedastic case, but also the level of heterogeneity in the variance.
Let $y_i | \mu_i \sim \mathcal{N}(\mu_i, \sigma_i)$ where $\mu_i$'s and $\sigma_i$'s are generated as follows:
\begin{align}
	\mu_i \overset{i.i.d}{\sim} w \delta_0 + (1-w) N(V, 1), \quad
	\sigma_i^2  \overset{i.i.d}{\sim} U(0.5, u)
\end{align}
where $\omega \in \{0.55, 0.65, \ldots, 0.95\}$, $V \in \{1, 1.5, 2, 2.5, 3\}$ and $u\in\{1, 1.5, 2, 2.5\}$.
For each setting, we set $n=1000$ and report the above-mentioned metrics across 100 Monte Carlo repetitions. We only show $u = 1.5$ and $V=2$ in the main text while the others are in the Appendix\footnote{The code to reproduce all the empirical results in a Github repository (\url{https://github.com/scsop/SNP}).}. 

We compare the following methods for posterior mean: 1) the generalized maximum likelihood Empirical Bayes estimator (GMLEB) of \cite{jiang2020general} using \cite{koenker2017rebayes}; 2) the group linear estimator by \cite{weinstein2018group}; 3) the semi-parametric monotonically constrained SURE estimator (XKB.SB) and the parametric SURE estimator (XKB.M) from \cite{xie2012sure}; 4) the Nonparametric Empirical Bayes Structural Tweedie (NEST) by \cite{banerjeenonparametric}; 5) the adaptive shrinkage (ash) by \cite{stephens2017false}\footnote{We obtain the SURE estimator by the code adapted by \cite{banerjeenonparametric} from \cite{weinstein2018group} and the HART estimator by \cite{banerjeenonparametric}.}.

For the class of posterior mode estimators, we compare 1) the parametric empirical Bayes median estimator (EBayesThresh) with Laplace tails \citep{johnstone_AS_2004}; 2) the SLOPE estimator \citep{bogdan2011asymptotic, su2016slope} with $q=0.1$; 3) the two-step Spike-and-Slab LASSO estimator \citep{rovckova2018bayesian}; 4) GMLEB posterior mode estimator \citep{koenker2017rebayes}.
% \footnote{ 
% We obtain EBayesThresh, SLOPE, and SSLASSO estimator using the EBayesThresh package by \cite{johnstone2005ebayesthresh}, the SLOPE package by \cite{rslope} and the SSLASSO \cite{rockova2019package} respectively.}.

For sparsity estimation, we can only compare methods mentioned above that are able to estimate sparsity.
Unfortunately, all the $f$-modeling approaches cannot provide sparsity estimation. Some methods estimate the sparsity as an intermediate step and we use that as the sparsity estimator.

For FDR control, we compare the empirical false discovery proportion $V/R$ and empirical power for various nominal FDR levels of 1) B\&H procedure \citep{benjamini1995controlling}; 2) Storey's procedure \citep{storey2003positive}; 3) Heterocedasticity-Adjusted Ranking and Thresholding procedure (HART) of \cite{fu2020heteroscedasticity}\footnote{We perform the Storey's procedure with \cite{dabney2010qvalue}. The code for HART is provided by the author.}.

\begin{figure*}
\centering
\includegraphics[width=\textwidth]{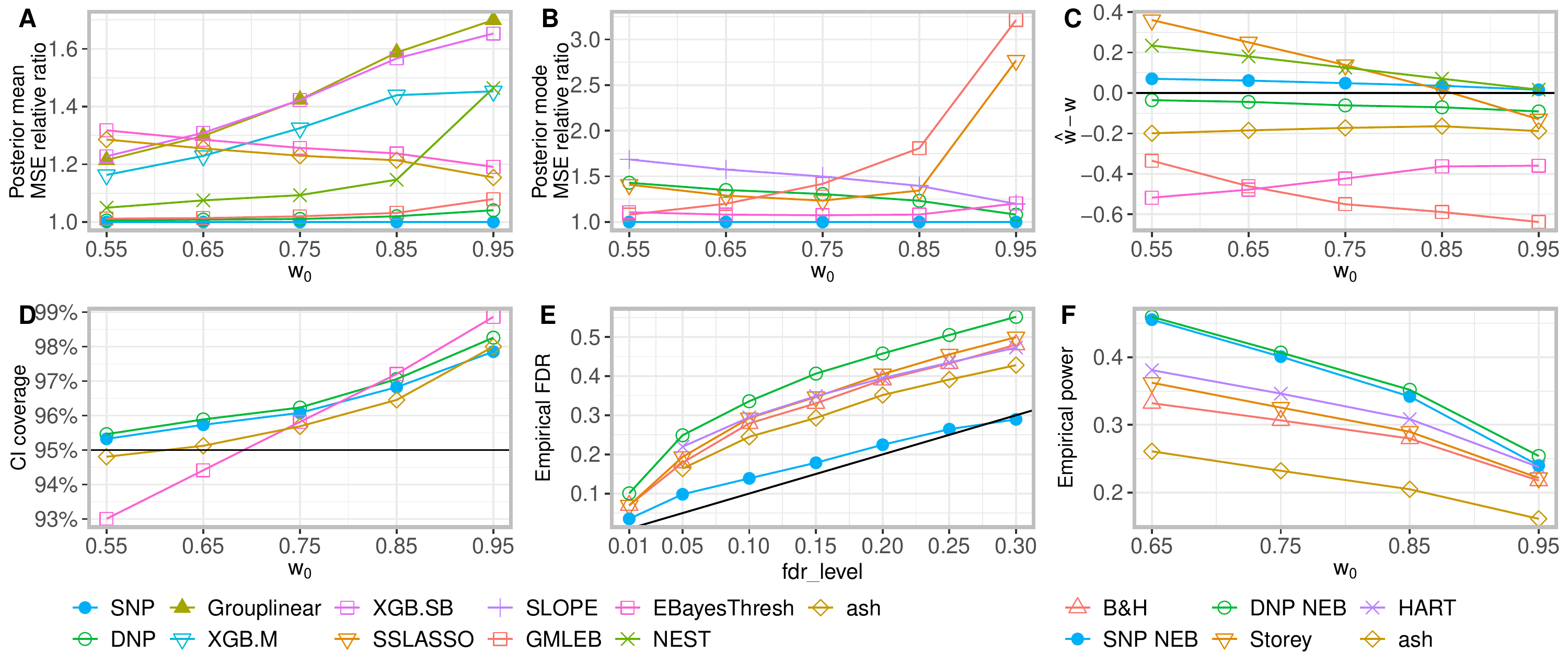}
\caption{\footnotesize{A) Relative mean Squared Error (MSE) of posterior mean
$n^{-1}\sum_{i=1} ^{n}(\hat{\mu}_i - \mu_i)^2$; B) Relative mean Squared Error (MSE) of posterior mode; C) Bias of sparsity estimation $\bias{\hat\omega} = \hat \omega - \omega$; D) Credible interval coverage; E) Empirical FDR; F) Empirical power controlling FDR at 0.05 level. Note that the relative ratio uses SNP as the base -- if the ratio is larger than 1, SNP performs better than the competing estimator.}}
\label{plot:sim}
% \vspace{-.1in}
\end{figure*}

Figure \ref{plot:sim} shows the robust performance of DNP and SNP in point estimation, sparsity estimation, and credible interval coverage at different sparsity levels. In addition, SNP-OPT controls FDR at various  levels and increases power as expected. In the Appendix, we show that it is also the case in general for various sparse, signal strength and heteroscedastic settings. 
Comparing DNP and SNP, the spike component of SNP demonstrates its power to capture the sparsity. DNP tends to underestimate sparsity, leading to an underestimate in the posterior probability of being zero and thus DNP-OPT is over-confident in rejecting hypotheses. 
%The adaptivity of the spike component is particularly attractive. 
Detailed results in tables are in the Appendix.

%%%%%%%%%%%%%%%%%%%%%%%%%%%%%%

\paragraph{Gene expression data}
We now apply our procedure on a prostrate cancer microarray study \citep{singh2002gene}\footnote{The microarray data is from the \textit{sda} package on CRAN.}. 
The data consists of $N=6,033$ genes measured on $n = 102$ subjects, among which $n_1 = 50$ healthy controls and $n_2 = 52$ prostate cancer patients. The goal is to identify the genes that help predict prostate cancer.
% and to predict whether or not a person has prostate cancer given his/her microarray.

\begin{figure*}
%   \begin{minipage}[c]{0.67\textwidth}
	\includegraphics[width=\textwidth]{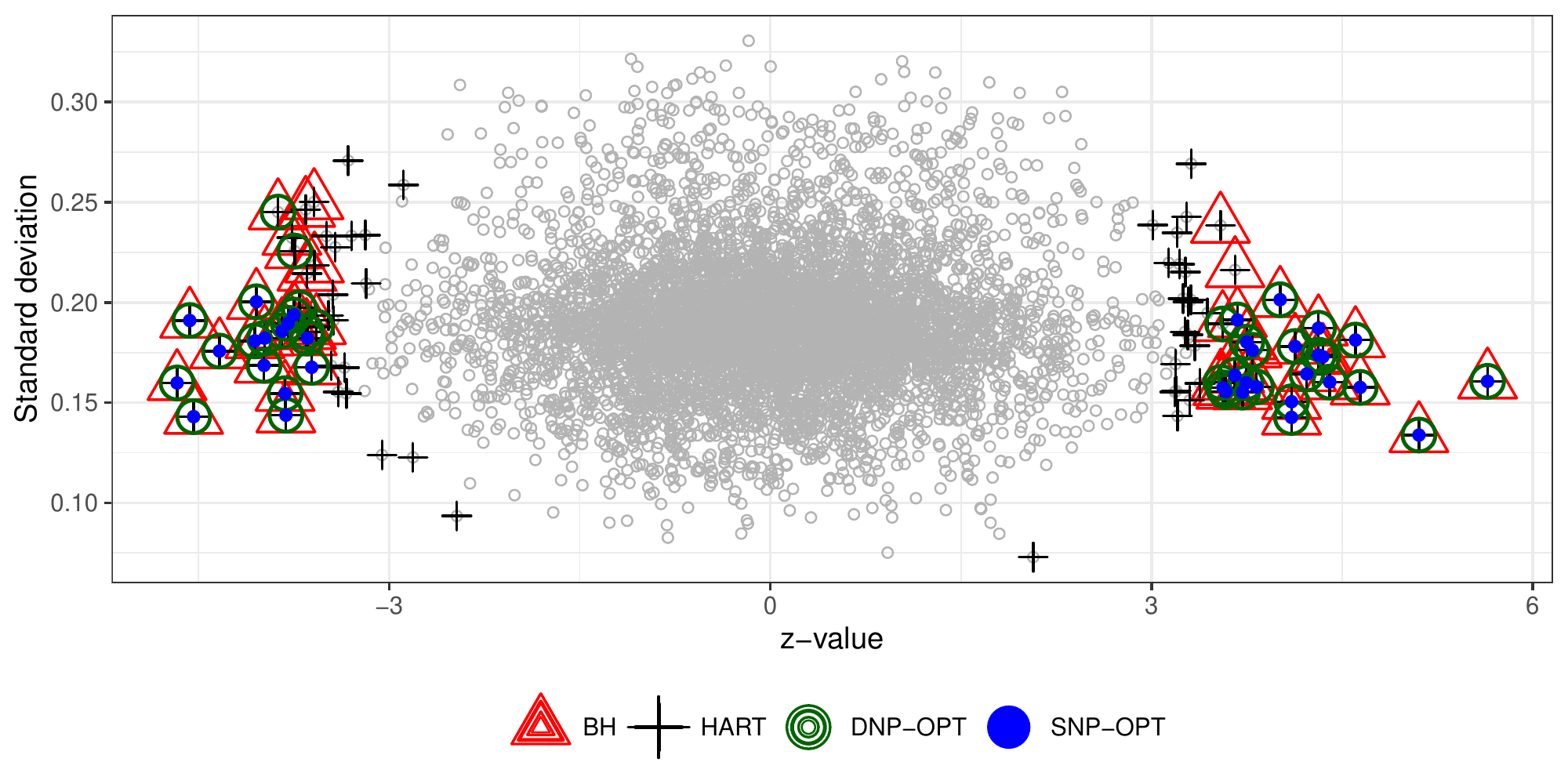}
%   \end{minipage}\hfill
%   \begin{minipage}[c]{0.33\textwidth}
    \caption{\footnotesize{The scatter plot of $Z$ vs $\sigma$. The red triangles ($\triangle$) label the 51 discoveries by the BH procedure. 
    The black plus ($+$) labels the 89 discoveries by HART. 
    The green circle ($\bigcirc$) labels the 44 discoveries by DNP-OPT. 
    % The purple cross ($\times$) labels the 29 discoveries by HART jackknifed procedure. 
    The blue solid circle ($\bullet$) label the 37 discoveries by SNP-OPT. 
    % The yellow solid triangle ($\blacktriangle$) label the 53 discoveries by Storey's procedure. 
    All the procedures target FDR level at 0.05.}
    } \label{fig:snp-bh}
%   \end{minipage}
%   \vspace{-.3in}
\end{figure*}

Let $y_i = \bar{x}_{i,2} - \bar{x}_{i,1}$ be the mean difference in gene $i$ expression between the cancer patients versus healthy subjects 
and $\hat{\sigma}^2_i = \frac{1}{n-2} \left[(n_2-1) s^{2}_{i,1} + (n_1-1) s^{2}_{i,2} \right]\left(\frac{1}{n_1} + \frac{1}{n_2}\right)$ be the pooled variance.
where $\bar{x}_{i,1}$ and $\bar{x}_{i,2}$ are the average gene expression of the healthy subjects and the cancer patients respectively; $s_{i,1} $ and $s_{i,2}$ are the standard error of the two groups. Define the $z$-value as $z_i = y_i/\hat\sigma_i$. 
Figure \ref{fig:snp-bh} shows the scatter plot of $z_i$ vs $\hat\sigma_i$.

We apply BH, HART as well as our SNP-OPT and DNP-OPT to the microarray data. 
All procedures target to control FDR at level 0.05.
For SNP-OPT and DNP-OPT, we simply plug in the differential difference and the pooled standard deviation estimate to our procedures. 
For BH, we will use the $p$-values converted from the $z$-value as 
$p_i = 2\Phi(-|z_i|)$ where $\Phi$ is the CDF of the standard normal distribution.
HART estimates the sparsity level using \cite{jin2007estimating} and provides  an optional jackknifed procedure to estimate the marginal density.

BH and HART reject more hypotheses than SNP-OPT and DNP-OPT.
Note that both BH and HART tend to be over-confident in the simulation.
We cannot say much since the ground truth is unknown. However, we should focus on the shape of the rejection region. As in Figure \ref{fig:snp-bh}, the rejection regions of DNP-OPT and SNP-OPT  depend on both $z_i$ and $\sigma_i$ -- hypotheses correspond to large $\sigma_i$ tend not to get rejected. 
We further compare with Storey's procedure and a variant of HART and adjust the empirical null as suggested in \cite{efron2004large} in the Appendix.
% Comparing with the simulation, BH, Storey and HART are overconfident.
% \footnote{(Indeed if we adjust the empirical null as suggested in \cite{efron2004large} to correct for the distortion using theoretical null $N(0,1)$, they reject fewer  hypotheses than SNP-OPT and DNP-OPT).}

%%%%%%%%%%%%%%%%%%%%%%%%%%%%%%%%%%%%%%%%%%%%%%%%%%%%%%%%%%%%
%%%%%%%%%%%%%%%%%%%%%%%%%%%%%%%%%%%%%%%%%%%%%%%%%%%%%%%%%%%%
\section{Summary and Discussion}
\label{sec:discussion}
%%%%%%%%%%%%%%%%%%%%%%%%%%%%%%%%%%%%%%%%%%%%%%%%%%%%%%%%%%%%
%%%%%%%%%%%%%%%%%%%%%%%%%%%%%%%%%%%%%%%%%%%%%%%%%%%%%%%%%%%%

In this paper, we propose a novel Spike-and-Nonparametric mixture prior (SNP) to tackle the high-dimensional sparse and heteroscedastic signal recovery problem. We develop an EM algorithm to estimate the unknown prior and derive the posterior. With the posterior, we are able to provide both point estimates with shrinkage/thresholding property and credible intervals for uncertainty quantification. In addition, our method estimates sparsity well, more accurate than the state-of-the-art methods. Based on our posterior, we also propose an optimal multiple testing procedure controlling FDR while achieving higher power. 

So far we only discuss the independent settings. However, our method can be extended to correlated $\y$ using a two-stage procedure -- we first estimate the prior ignoring correlations; and then we apply Markov chain Monte Carlo (MCMC) with Gibbs samplers to obtain the posterior.
%
%\begin{ack}
%Use unnumbered first level headings for the acknowledgments. All acknowledgments
%go at the end of the paper before the list of references. Moreover, you are required to declare
%funding (financial activities supporting the submitted work) and competing interests (related financial activities outside the submitted work).
%More information about this disclosure can be found at: \url{https://neurips.cc/Conferences/2021/PaperInformation/FundingDisclosure}.
%
%Do {\bf not} include this section in the anonymized submission, only in the final paper. You can use the \texttt{ack} environment provided in the style file to automatically hide this section in the anonymized submission.
%\end{ack}

%%%%%%%%%%%%%%%%%%%%%%%%%%%%%%%%%%%%%%%%%%%%%%%%%%%%%%%%%%%%

\bibliographystyle{plainnat}
\bibliography{biblio}

%%%%%%%%%%%%%%%%%%%%%%%%%%%%%%%%%%%%%%%%%%%%%%%%%%%%%%%%%%%%

%%%%%%%%%%%%%%%%%%%%%%%%%%%%%%%%%%%%%%%%%%%%%%%%%%%%%%%%%%%%

\appendix

\section{Appendix}

Please find the Appendix in the supplemental material.
%%%%%%%%%%%%%%%%%%%%%%%%%%%%%%%%%%%%%%%%%%%%%%%%%%%%%%%%%%%%

\end{document}

\onecolumn

\aistatstitle{Supplementary Materials for Nonparametric Empirical Bayes Estimation and Testing for Sparse and Heteroscedastic Signals}

\aistatsauthor{ Junhui Cai \And Xu Han \And  Ya'acov Ritov \And Linda Zhao }

\aistatsaddress{ University of Pennsylvania \And Temple University \And University of Michigan \And University of Pennsylvania } 
% ]

% \maketitle

% \appendix
\pagebreak

This supplementary material provides detailed derivation, proofs and more empirical results 
that  are deferred from the main text.
The following is organized as follows.
Section \ref{app:sim} provides more simulation results.
Section \ref{app:gene} further compares SNP-OPT and DNP-OPT with the Storey's procedure and HART jacknife for the gene expression data case study. We further adjust the empirical null as suggested in \cite{efron2004large} for BH, Storey, and HART and compare with our proposed procedures.
%Proofs of  Theorem \ref{Main-thm:dnp-em} and Proposition \ref{Main-prop1} are in Section \ref{app:proof}.
Proofs of  Theorem 1 and Proposition 1 are in Section \ref{app:proof}.
The simulation results presented in tables are in Section \ref{app:sim-tbl}.

%%%%%%%%%%%%%%%%%%%%%%%%%%%%%%%%%%%%%%%%%%%%%%%%%%%%%%%%%%%%
%%%%%%%%%%%%%%%%%%%%%%%%%%%%%%%%%%%%%%%%%%%%%%%%%%%%%%%%%%%%
\section{Simulation}
\label{app:sim}
%%%%%%%%%%%%%%%%%%%%%%%%%%%%%%%%%%%%%%%%%%%%%%%%%%%%%%%%%%%%
%%%%%%%%%%%%%%%%%%%%%%%%%%%%%%%%%%%%%%%%%%%%%%%%%%%%%%%%%%%%

The goal of our simulation is to investigate the adaptivity of DNP and SNP to different levels of signal strength, sparsity level, and heteroscedasticity.
%Therefore, we vary not only the signal strength $V$ and the sparsity level $w$ as in the homoscedastic case, but also the level of heterogeneity in the variance.
To be specific, $y_i | \mu_i \sim \mathcal{N}(\mu_i, \sigma_i)$ where $\mu_i$'s and $\sigma_i$'s are generated as follows:
\begin{align}
	\mu_i \overset{i.i.d}{\sim} w \delta_0 + (1-w) N(V, 1), \quad
	\sigma_i^2  \overset{i.i.d}{\sim} U(0.5, u)
\end{align}
where $\omega \in \{0.55, 0.65, \ldots, 0.95\}$, $V \in \{1, 1.5, 2, 2.5, 3\}$ and $u\in\{1, 1.5, 2, 2.5\}$.
Note that $V \in \{1,1.5,2\}$
For each setting, we set $n=1000$ and report the above-mentioned metrics across 100 Monte Carlo repetitions.

We compare the following metrics: 
\begin{enumerate}[leftmargin=*]
	\item Relative mean Squared Error (MSE) of posterior mean
$n^{-1}\sum_{i=1} ^{n}(\hat{\mu}_i - \mu_i)^2$;
\item Relative mean Squared Error (MSE) of posterior mode;
\item Bias of sparsity estimation $\bias{\hat\omega} = \hat \omega - \omega$;
\item Credible interval coverage $n^{-1}\sum_{i=1} ^{n} \bm{1}\{\mu \in \widehat{CI}\}$;
\item Empirical FDR controlling FDR at different nominal levels;
\item Empirical power controlling FDR at 0.05 level. 
\end{enumerate}
Note that the relative ratio uses SNP as the base, i.e., the relative ratio is the ratio of the metric for any competing estimator to that of SNP. If the ratio is larger than 1, SNP performs better than the competing estimator.

%Figure \ref{Main-plot:sim} 
Figure 1 in the main manuscript compares different methods in terms of different metrics with $u=1.5$ and $V=2$. We provide the tables corresponding to 
%Figure \ref{Main-plot:sim} 
Figure 1
for detailed comparisons. Table \ref{tab:tbl:MSE-u1.5-V2} to \ref{tab:tbl:emp-power-u1.5-fdr_level0.05} shows the MSE of posterior mean, the relative MSE of posterior mean, the MSE of posterior mode, the relative MSE of posterior mode, the bias of the sparsity estimate, the coverage and average length of credible interval, the empirical FDR with $\omega = 0.95$ and the empirical power controlling FDR level at 0.05 respectively.

% \begin{figure}
% \centering
% \includegraphics[width=\linewidth]{\localpath /simulation/plot/nips_sim}
% \caption{\footnotesize{A) Relative mean Squared Error (MSE) of posterior mean
% $n^{-1}\sum_{i=1} ^{n}(\hat{\mu}_i - \mu_i)^2$; B) Relative mean Squared Error (MSE) of posterior mode; C) Bias of sparsity estimation $\bias{\hat\omega} = \hat \omega - \omega$; D) Credible interval coverage; E) Empirical FDR; F) Empirical power controlling FDR at 0.05 level. Note that the relative ratio uses SNP as the base -- if the ratio is larger than 1, SNP performs better than the competing estimator.}}
% \label{Main-app-plot:sim}
% \vspace{-.1in}
% \end{figure}

Figure \ref{plot:het-mse} to \ref{plot:het-emp-power-w95} show more detailed comparisons varying $u$ and $V$. In Section \ref{app:sim-tbl}, we provide all the information in the table format with the best performers set in bold type.

Figure \ref{plot:het-mse} reports the relative MSE of posterior mean varying the signal strength $V$, sparsity level $w$ and variance heterogeneity $u$. The columns correspond to different signal strength $V$ and the rows are across different variance heterogeneity $u$. For each plot, the x-axis is the sparsity level $w_0$ and the y-axis the ratio of the MSE of competing estimator to the MSE of SNP. 
We compare with  
1) the parametric empirical Bayes mean estimator (EBayesThresh) with Laplace as the slab of \cite{johnstone_AS_2004}; 
2) the generalized maximum likelihood Empirical Bayes estimator (GMLEB) of \cite{jiang2020general} using \cite{koenker2017rebayes}; 
3) the group linear estimator by \cite{weinstein2018group}; 
4) the semi-parametric monotonically constrained SURE estimator (XKB.SB) and the parametric SURE estimator (XKB.M) from \cite{xie2012sure}; 
5) the Nonparametric Empirical Bayes Structural Tweedie (NEST) by \cite{banerjeenonparametric};
6) adaptive thresholding by \cite{stephens2017false}. 
Note that NEST is designed for unknown variance. 
We can see two clusters in terms of performance, one of the parametric methods and one of the nonparametric. The nonparametric methods perform better than the parametric counterparts. When the signal $V$ is strong, the advantages of the nonparametric methods are even larger.
In general, SNP and DNP perform better than the others. The advantages is more pronounced as the sparsity level $w_0$ increases since the other methods are not specially designed for sparse data. The closest competitor with SNP and DNP is GMLEB which adapts $g$-modeling as well but does not use an EM algorithm. NEST is also comparable when the heterogeneity $u$ is relatively small. 

\def\uu{1.5}
\hettbl{MSE_V2}{\uu}
\hettbl{MSE_ratio_V2}{\uu}
\hettbl{MSE_mode_V2}{\uu}
\hettbl{MSE_mode_ratio_V2}{\uu}
\hettbl{w_diff_V2}{\uu}
\hettbl{ci_coverage_V2}{\uu}
\hettbl{ci_len_V2}{\uu}
\hettbl{emp_fdr_w0.95_V2}{\uu}
\hettbl{emp_power_fdr_level0.05}{\uu}

%============================
\plotfig{\hetpath}{MSE}{Posterior mean MSE relative ratio.}{plot:het-mse}

%============================
\plotfig{\hetpath}{MSE_mode}{Posterior mode MSE relative ratio.}{plot:het-mode}

Figure \ref{plot:het-mode} reports the performance of the posterior mode estimators varying the signal strength $V$, sparsity level $w$ and variance heterogeneity $u$. 
We compare 1) the parametric empirical Bayes mode estimator (EBayesThresh) with Laplace as the slab of \cite{johnstone_AS_2004}; 2) the SLOPE estimator of \cite{bogdan2011asymptotic, su2016slope} with $q=0.1$; 3) the two-step Spike-and-Slab LASSO estimator of \cite{rovckova2018bayesian}; 4) GMLEB posterior mode estimator. 
SNP is robust across different settings; while DNP can be less competitive since it tends to underestimate the sparsity. Comparing SNP and DNP shows the advantages of the Laplacian spike capturing sparsity. Surprisingly, our closest competitor in posterior mean estimator, GMLEB, is quite off using its posterior mode estimator. This may due to the convex relaxation \cite{koenker2017rebayes}.

Figure \ref{plot:het-w} shows the performance of sparsity estimation $\hat\omega$ varying the signal strength $V$, sparsity level $w$ and variance heterogeneity $u$. 
We can only compare methods mentioned above that are able to estimate sparsity. Unfortunately, all the $f$-modeling approaches cannot provide sparsity estimation. 
Therefore, we only compare DNP and SNP with GMLEB, SSLASSO, EBayesThresh, HART and ash. Note that HART uses \cite{jin2007estimating} with a theoretical null $N(0,1)$ to estimate the sparsity. 
SNP and DNP estimate the sparsity level with high accuracy, while DNP  underestimates the sparsity when the heterogeneity $u$ is low compared to SNP. GMLEB, EBayesThresh and ash underestimate the sparsity level, while HART and SSLASSO tend to overestimate. The overestimating behavior of SSLASSO agrees with Theorem 4.2 of \cite{rovckova2018bayesian}.

%============================
\plotfig{\hetpath}{w_diff}{Bias of the sparsity.}{plot:het-w}

%============================
\plotfig{\hetpath}{ci}{Coverage and width of the credible interval.}{plot:het-ci}

As a bonus of the Bayesian mechanism, we are able to provide uncertainty quantification in addition to point estimate. We construct the 95\% equal-tailed credible interval from the posterior distribution. Figure \ref{plot:het-ci} shows the empirical coverage and the average width of the credible interval across $w$ and $V$, comparing with EBayesThresh and ash. Most credible intervals are overshoot while the widths of the interval are acceptable. The credible intervals are below nominal coverage when signal is weak and the heterogeneity of noise is large. In general, the average length of the credible interval by ash is wider than that of DNP and SNP. The credible interval by EBayesThresh is below the nominal coverage, especially with high signal and heterogeneity levels and is wider than the others.

Figure \ref{plot:het-fdr}, \ref{plot:het-fdr-w95}, \ref{plot:het-emp-power-u15} and \ref{plot:het-emp-power-w95} show the performance of the multiple testing procedure. Each plot is at different signal strength $V$ and different heterogeneity level $u$ or sparsity level $w$. The x-axis is the FDR control level ($\alpha$) and the y-axis is the bias of the average of the empirical ratio $V/R$ and the empirical power across 100 runs. We compare SNP-OPT and DNP-OPT with the original linear step-up \cite{benjamini1995controlling} procedure, the pFDR of \cite{storey2002direct}, the adaptive threshold (ash) of \cite{stephens2017false}, and the HART procedure of \cite{fu2020heteroscedasticity}.

Figure \ref{plot:het-fdr} shows the empirical FDR varying the signal strength $V$ and sparsity level $w$ when the heterogeneity $u=1.5$ across different nominal levels; while Figure \ref{plot:het-fdr-w95} fixes the sparsity level $w=0.95$ and varies the signal strength $V$ and the heterogeneity level. 
SNP-OPT controls false discovery rate at the desired nominal level $\alpha$, while others are overconfident and reject too many hypotheses in most of the cases. 
When the sparsity level is low, B\&H procedure is conservative as expected. 
Storey's procedure is quite robust since it estimates the error rate of a predetermined rejection region (other than the sparse setting when $w = 0.95$).
It is instructive to compare DNP-OPT and SNP-OPT to see the benefits of the Laplacian spike. DNP-OPT tends to reject too many hypotheses, overshooting the nominal level $\alpha$. This is expected since DNP underestimates the sparsity level as in Figure \ref{plot:het-w}. As a result, the posterior probability of being zero $\hat{p}_{i}(y_i)$ is underestimated. Therefore, following our NEB-OPT procedure, DNP-OPT is over-confident in rejecting hypotheses. 

Similarly, Figure \ref{plot:het-emp-power-u15} shows the empirical power by fixing the heterogeneity level at $u=1.5$ and varying the signal strength $V$ and sparsity level $w$ across different nominal FDR levels; while Figure \ref{plot:het-emp-power-w95} fixes the sparsity level $w=0.95$ and varies the signal strength $V$ and the heterogeneity level. 
SNP-OPT controls FDR at nominal levels and at the same time increases power in most settings.

%============================
\plotfig{\hetpath}{fdr}{FDR control at heterogeneity level $u = 1.5$.}{plot:het-fdr}

%============================
\plotfig{\hetpath}{fdr_w95}{FDR control at sparsity level $w =0.95$.}{plot:het-fdr-w95}

%============================
\plotfig{\hetpathII}{emp_power_u15}{Empirical power  at heterogeneity level $u = 1.5$.}{plot:het-emp-power-u15}

%============================
\plotfig{\hetpathII}{emp_power_w95}{Empirical power at sparsity level $w =0.95$. }{plot:het-emp-power-w95}

This concludes the simulation study. The nonparametric mixture prior is robust and versatile across various sparsity levels and signal strengths, especially for SNP. The multi-directional shrinkage property is particular desirable in the sparse set up where the noises are shrunk towards zero while the signals towards their corresponding centers. The multi-directional effect also reflects in the adaptive thresholding for the posterior mode estimator. In addition to point estimate, uncertainty quantification is readily provided from the Bayesian mechanism. The equal-tailed credible intervals show coverage at nominal level and are of reasonable widths. 

With a focus on sparse data, it is desirable to have an accurate sparsity estimate. SNP is able to estimate the sparsity level by nature, and it estimates the sparsity well. It had been troubling to the authors that DNP tends to underestimates the sparsity due to the dispersion around zero. The underestimate of DNP also compromises the performance of the multiple testing procedures. Remedies such as a wider gap between zero and the grid points around zero are not adaptive and rather artificial. 
SNP, on the other hand, replaces the point-mass at zero with a Laplacian spike to handle the sparsity. The adaptivity of the spike component is particularly attractive. 

The proposed multiple testing procedures, NEB-OPT, control FDR at nominal levels in different settings and achieve higher power. 

%%%%%%%%%%%%%%%%%%%%%%%%%%%%%%%%%%%%%%%%%%%%%%%%%%%%%%%%%%%%

%%%%%%%%%%%%%%%%%%%%%%%%%%%%%%%%%%%%%%%%%%%%%%%%%%%%%%%%%%%%
%%%%%%%%%%%%%%%%%%%%%%%%%%%%%%%%%%%%%%%%%%%%%%%%%%%%%%%%%%%%
\section{Gene expression data}
\label{app:gene}
%%%%%%%%%%%%%%%%%%%%%%%%%%%%%%%%%%%%%%%%%%%%%%%%%%%%%%%%%%%%
%%%%%%%%%%%%%%%%%%%%%%%%%%%%%%%%%%%%%%%%%%%%%%%%%%%%%%%%%%%%

\begin{figure}
	\includegraphics[width=\textwidth]{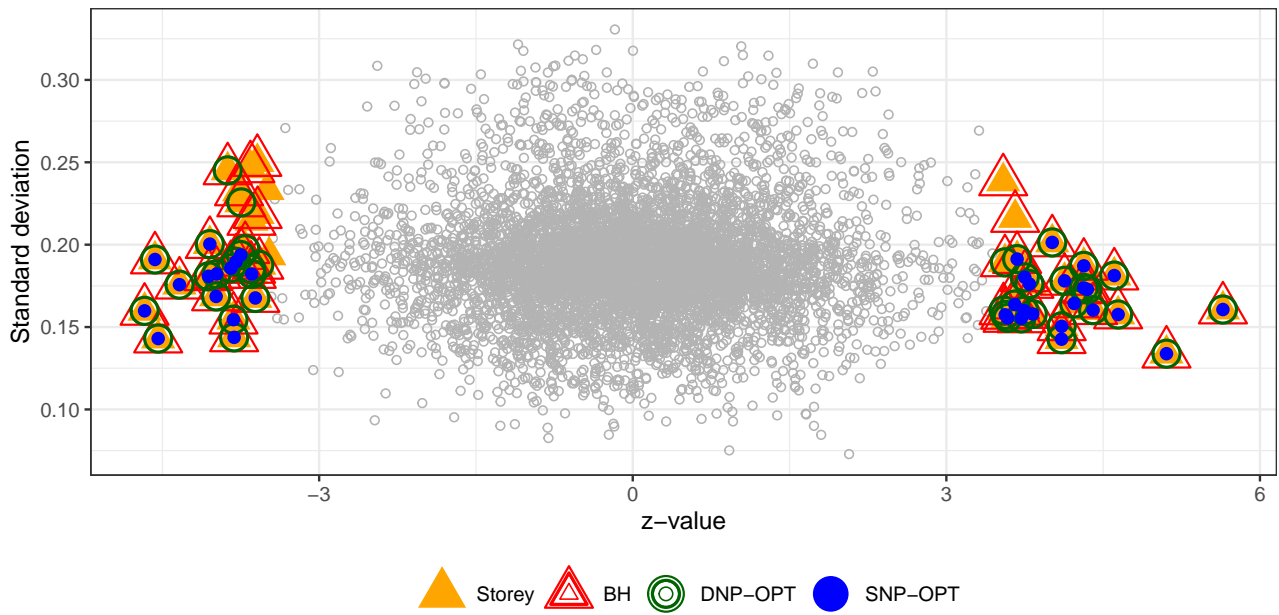}
    \caption{\footnotesize{The scatter plot of $Z$ vs $\sigma$. The red triangles ($\triangle$) label the 51 discoveries by the BH procedure. The green circle ($\bigcirc$) labels the 44 discoveries by DNP-OPT. 
The blue solid circle ($\bullet$) label the 37 discoveries by SNP-OPT. The yellow solid triangle ($\blacktriangle$) label the 53 discoveries by Storey's procedure. All the procedures target FDR level at 0.05.}
    } \label{Main-app-fig:snp-dnp-bh}
\end{figure}

We first compare different methods as in 
%Figure \ref{Main-fig:snp-bh}
Figure 2
 of the main manuscript but in separate plots for clearer comparison. Figure \ref{Main-app-fig:snp-dnp-bh} compares DNP and SNP with the classic the BH and Storey's procedure. 
%Figure \ref{Main-app-fig:snp-hart} 
Figure 2
compares SNP with the state-of-the-art HART and its jackknifed variant.

\begin{figure}
	\includegraphics[width=\textwidth]{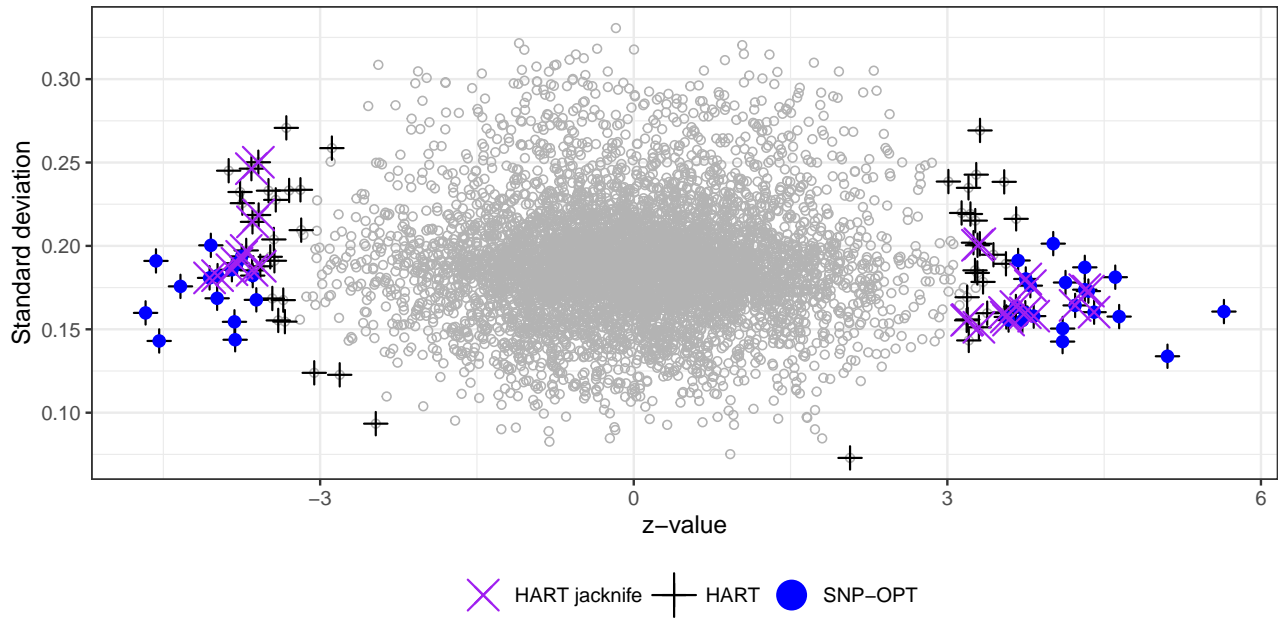}
    \caption{\footnotesize{The scatter plot of $Z$ vs $\sigma$. 
    The black plus ($+$) labels the 89 discoveries by HART. The purple cross ($\times$) labels the 29 discoveries by HART jackknifed procedure. The blue solid circle ($\bullet$) label the 37 discoveries by SNP-OPT. All the procedures target FDR level at 0.05.}
    } \label{Main-app-fig:snp-hart}
\end{figure}

\begin{figure}
\centering
\subcaptionbox{Histogram of the unadjusted $p$-value \label{fig:pval}}{\includegraphics[width=0.45\textwidth]{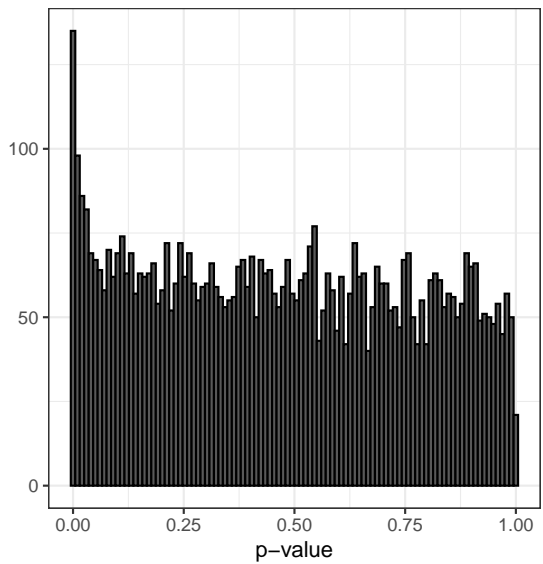}}%
\hfill
\subcaptionbox{Histogram of the adjusted $p$-value\label{fig:pval_adjusted}}{\includegraphics[width=0.45\textwidth]{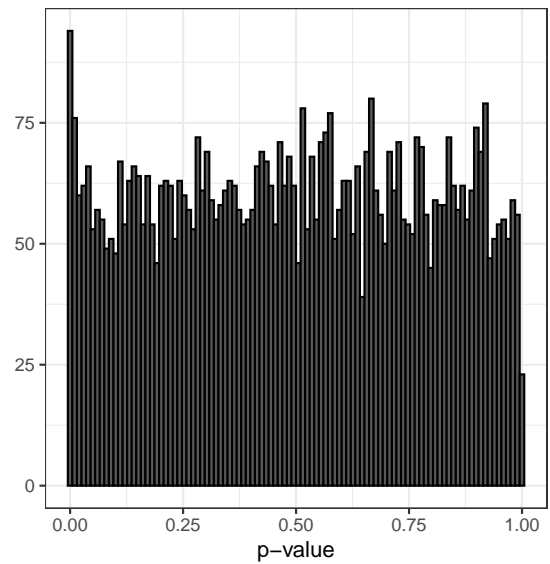}}%
\caption{Figure (A) shows the histogram of the unadjusted $p$-value. Figure (B) shows the histogram of the adjusted $p$-value $p'_i$'s estimated from the empirical null $N(0, 1.09^2)$. }
\end{figure}

As argued in \cite{efron2004large}, a small deviation from the theoretical null $N(0,1)$ will distort the FDR analysis, resulting in too many inappropriate rejections as in Figure \ref{Main-fig:snp-bh}. In order to compare our methods and other methods thoroughly, we adopt the empirical null approach in \cite{efron2004large} to first estimate the empirical null. The estimated empirical null turns out to be $N(0,1.09^2)$. We then obtain the new $p$-value by converting the $z$-value as $p'_i = 2\Phi'(-|Z_i|)$ where $\Phi'$ is the CDF of a $N(0,1.09^2)$ variable. We can see that comparing Figure \ref{fig:pval} and \ref{fig:pval_adjusted}, the histogram of $p$-value estimated from the empirical null is closer to uniform compared to that estimated from the theoretical null $N(0,1)$.

% latex table generated in R 4.0.3 by xtable 1.8-4 package
% Wed May 19 14:54:00 2021
\begin{table}[ht]
\centering
\begin{tabular}{cccc}
  \hline
Storey & HART & DNP-OPT & SNP-OPT \\ 
  \hline
0.93 & 0.99 & 0.91 & 0.96 \\ 
   \hline
\end{tabular}
\caption{Sparsity level estimation} 
\label{tbl:microarray-w-est-adjusted}
\end{table}

We are now ready to apply BH, Storey, HART as well as our SNP-OPT and DNP-OPT to the microarray data. 
All procedures target to control FDR at level 0.05.
For SNP-OPT and DNP-OPT, we simply plug in the differential difference and the pooled estimate of the standard deviation to our procedures. 
For the others, we will use the empirical null and the adjusted $p$-values $p'_i$.
Both BH and Storey use the $p$-value $p'_i$ estimated from the empirical null.
HART estimates the sparsity level using Jin-Cai's method with the empirical null $N(0,1.09^2)$, following the procedure as in \cite{fu2020heteroscedasticity}, and adopts a jacknifed procedure to estimate the marginal density. Since SNP also provides an estimate to the sparsity, we can plug in to the HART procedure as an alternative approach to estimate the non-null proportion.

% latex table generated in R 4.0.3 by xtable 1.8-4 package
% Wed May 19 14:53:59 2021
\begin{table}[ht]
\centering
\begin{tabular}{ccccccc}
  \hline
BH & Storey & HART jk & HART jk + SNP & DNP-OPT & SNP-OPT & SNP Mode \\ 
  \hline
13 & 13 & 19 & 29 & 44 & 37 & 59 \\ 
  (0.22\%) & (0.22\%) & (0.31\%) & (0.48\%) & (0.73\%) & (0.61\%) & (0.98\%) \\ 
   \hline
\end{tabular}
\caption{Number and proportion of discoveries controlling FDR level at 0.05} 
\label{tbl:microarray-num-dis-adjusted}
\end{table}

\begin{figure}
\centering
\includegraphics[width=0.95\textwidth]{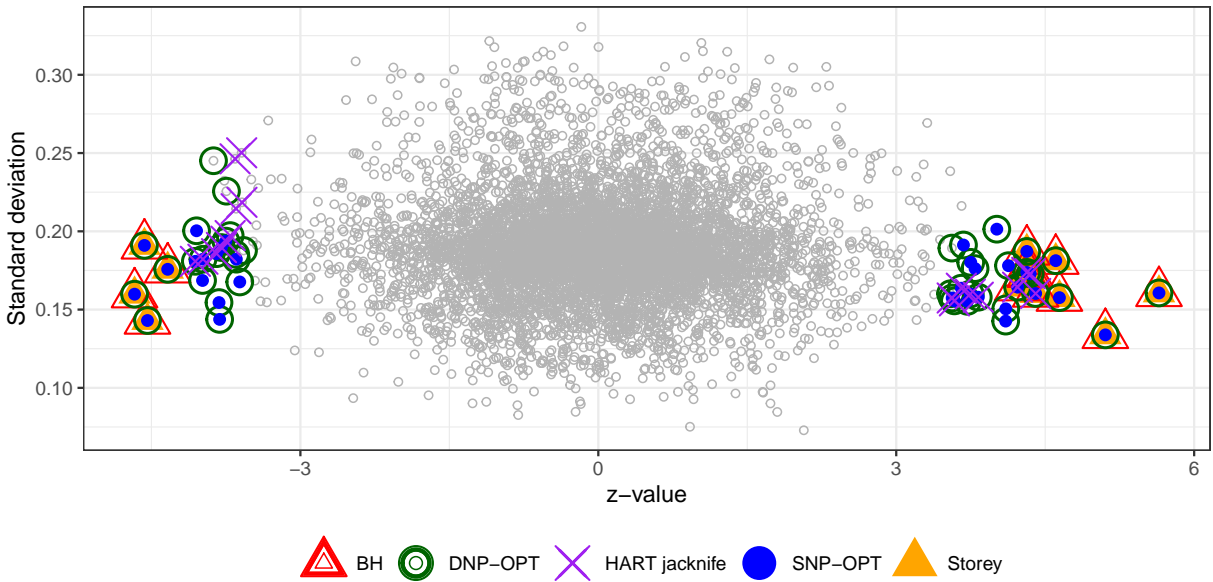}
\caption{The scatter plot of $Z$ vs $\sigma$. The red triangles ($\triangle$) label the 13 discoveries by the BH procedure. The green circle ($\bigcirc$) labels the 44 discoveries by DNP-OPT. The purple cross ($\times$) labels the 19 discoveries by HART jackknifed procedure. The blue solid circle ($\bullet$) label the 37 discoveries by SNP-OPT. The yellow solid triangle ($\blacktriangle$) label the 13 discoveries by Storey's procedure. All the procedures target FDR level at 0.05.}
\label{fig:snp-bh}
\end{figure}

Table \ref{tbl:microarray-num-dis-adjusted} shows the number of discoveries and Figure \ref{fig:snp-bh} shows the discoveries on the $Z$ vs $\sigma$ scatter plot controlling FDR at 0.05 level. SNP-OPT and DNP-OPT reject more hypotheses than other methods. SNP-OPT rejects 37 hypotheses and DNP-OPT rejects 44. 
The BH procedure is the most conservative as expected, claiming 13 discoveries. Note that if we use the unadjusted $p$-value from the theoretical null, the BH procedure yield 51 discoveries, many of which could be over-confident false rejections as we see in the simulation. The Storey procedure obtains similar results. The HART procedure yields 19 discoveries while the HART variant using the sparsity estimated by SNP yields 29 discoveries. 

Since we do not know the ground truth, we again cannot claim much. However, it is insightful to compare the rejection regions as in Figure \ref{fig:snp-bh}. The reject regions for SNP-OPT and DNP-OPT depend on both $Z$ and $\sigma$. The dependency is more obvious for SNP-OPT -- SNP-OPT does not reject hypothesis that corresponds to large $\sigma_i$.

% latex table generated in R 4.0.3 by xtable 1.8-4 package
% Wed May 19 14:54:16 2021
\begin{table}[ht]
\centering
\begin{tabular}{cccc}
  \hline
Storey & HART & DNP-OPT & SNP-OPT \\ 
  \hline
0.93 & 0.99 & 0.91 & 0.96 \\ 
   \hline
\end{tabular}
\caption{Sparsity level estimation} 
\label{tbl:microarray-w-est}
\end{table}

Table \ref{tbl:microarray-w-est} shows the sparsity estimation. Storey's procedure estimates the sparsity at 0.93 while the Jin-Cai procedure \citep{jin2007estimating} used by HART estimates the sparsity at 0.99. SNP-OPT, which demonstrates the accurate sparsity estimation in the simulation, estimates 0.96, in between the two mentioned.

We also compare the posterior mode estimators. The posterior mode estimator of SNP produces 59 non-zero estimate although it does not provide guarantee for FDR control. On the other hand, SLOPE rejects 0 hypothesis when controlling FDR at 0.05 level.

%\begin{figure}
%\centering
%\includegraphics[width=0.95\textwidth]{\microarraypath{snp_bh}}
%\caption{The standardized mean differences of 6,033 gene expression between the healthy subjects and the cancer patients. The red bars are the 37 genes identified by the SNP-OPT procedure controlling FDR level at 0.05; the blue bars are the addition 14 genes (51 genes in total) identified by the BH controlling FDR level at 0.05.}
%\label{fig:snp-bh}
%\end{figure}

\begin{figure}
\centering
\includegraphics[width=0.95\textwidth]{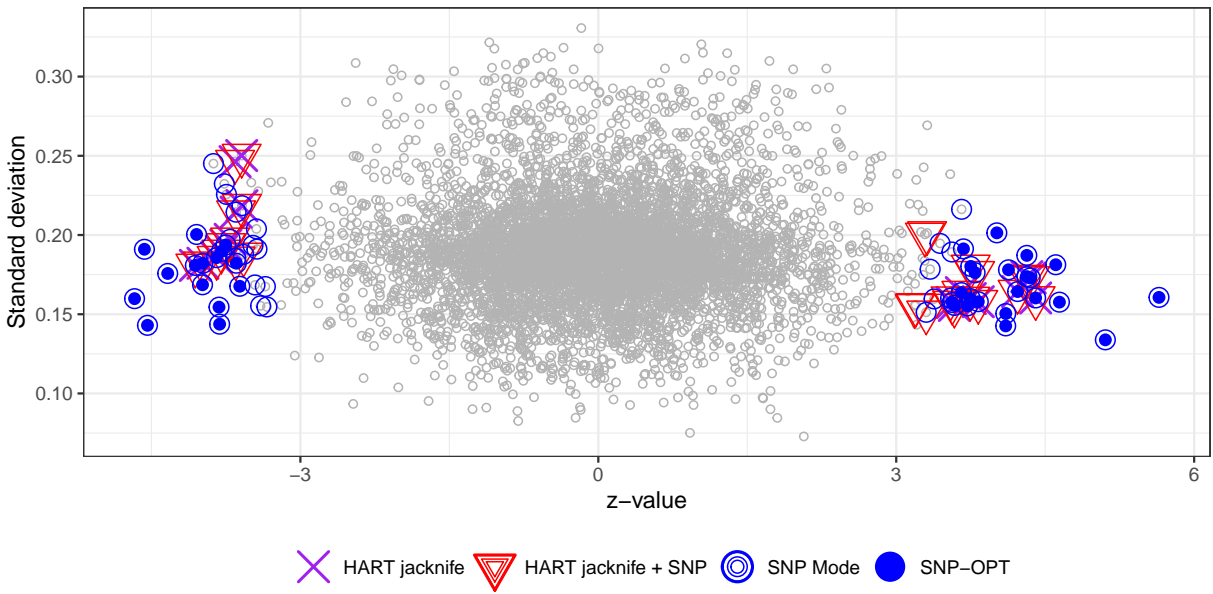}
\caption{The scatter plot of $Z$ vs $\sigma$. 
The purple cross ($\times$) labels the 19 discoveries by HART jackknifed procedure.
The red triangle ($\triangledown$) label the 29 discoveries by HART with SNP sparsity plug in. The blue circle ($\bigcirc$) labels the 59 non-zero estimate from the posterior mode estimator of SNP.
 The blue solid circle ($\bullet$) label the 37 discoveries by SNP-OPT. All the multiple testing procedures target FDR level at 0.05.}
\label{fig:snp-hart}
\end{figure}

\begin{figure}
\centering
\includegraphics[width=0.95\textwidth]{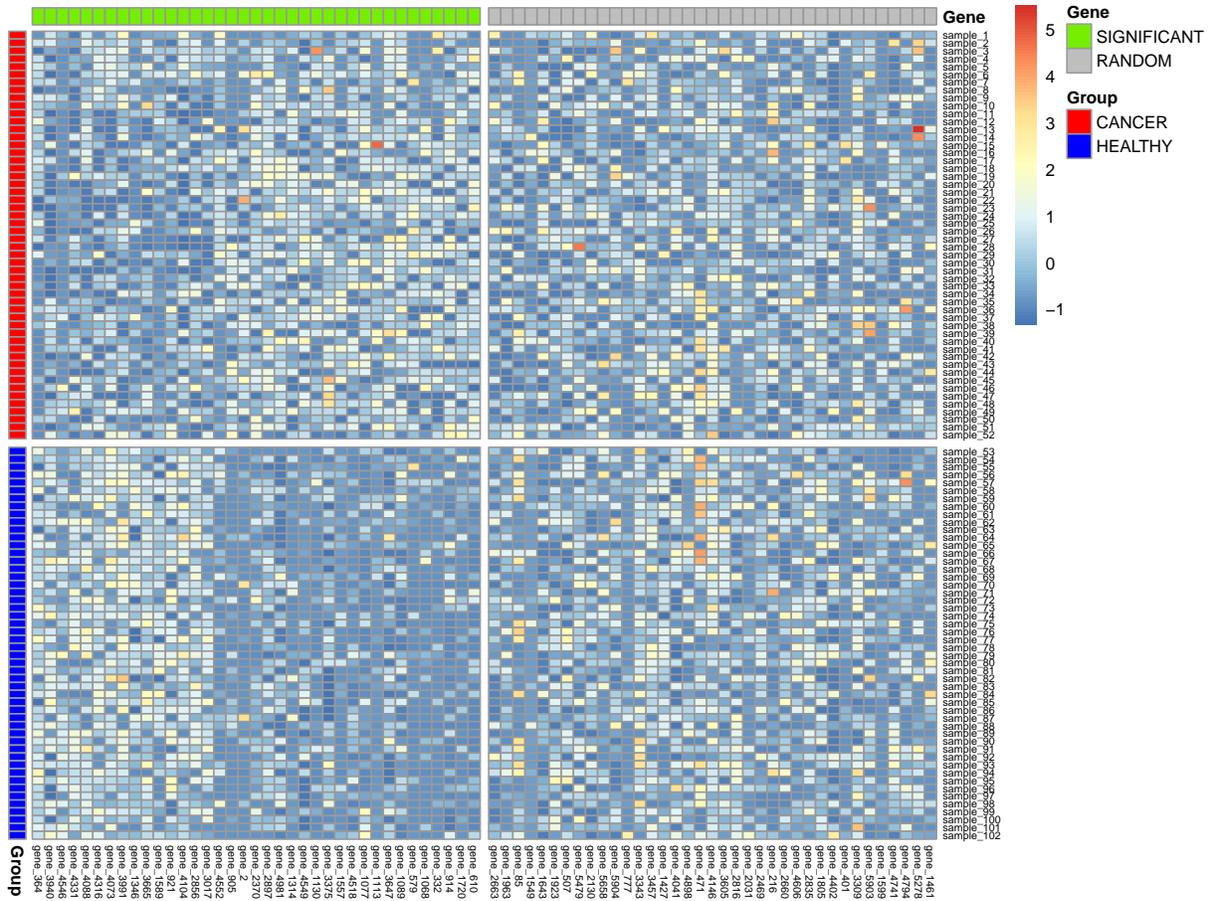}
\caption{The heatmap of the discovery genes by SNP-OPT vs non-discovery genes among the cancer patients and healthy subjects. The two vertical panels are the significant group (discoveries) with the header labelled in green and the randomly selected group from the non-discoveries with the header labelled in grey. The two horizontal panels are the cancer patients with the side bar labelled in red and the healthy group with the side bar lablled in blue.}
\label{fig:heatmap}
\end{figure}

Lastly, Figure \ref{fig:heatmap} shows the heatmap of the discovery genes by SNP-OPT vs non-discovery genes among the cancer patients and healthy subjects. Comparing the differential expression of the significant genes (discoveries) on the left and the randomly picked from the non-discoveries on the right, we see that the difference in expression level among the the cancer patients and healthy subjects is distinguishable from the significant genes, while the difference in the non-discovery group is less obvious. 

%%%%%%%%%%%%%%%%%%%%%%%%%%%%%%%%%%%%%%%%%%%%%%%%%%%%%%%%%%%%
%%%%%%%%%%%%%%%%%%%%%%%%%%%%%%%%%%%%%%%%%%%%%%%%%%%%%%%%%%%%
\section{Proofs}
\label{app:proof}
%%%%%%%%%%%%%%%%%%%%%%%%%%%%%%%%%%%%%%%%%%%%%%%%%%%%%%%%%%%%
%%%%%%%%%%%%%%%%%%%%%%%%%%%%%%%%%%%%%%%%%%%%%%%%%%%%%%%%%%%%

%%%%%%%%%%%%%%%%%%%%%%%%%%%%%%%%%%%%%%%%%%%%%%%%%%%%%%%%%%%%
%\subsection{Proof of Theorem \ref{Main-thm:dnp-em}}
\subsection{Proof of Theorem 1}

 \begin{lemma}
 	Suppose we have an estimate of $	\piold$ in round $t$. Define 
 	\begin{equation}
 		\label{cond-z}
 		\piold_{j,i} = \frac{p(y_i~|~ \tau_j)~ \piold_j}{\sum_{m=1}^M p(y_i ~|~ \tau_m)~ \piold_m}.
	\end{equation}
The EM algorithm update $\bm\pi$ by
\begin{equation}
	\label{em-sol}
	\pinew_j  = \frac{1}{n}\sum_{i=1}^{n}\piold_{j,i}.
\end{equation}
 \end{lemma}
 
 \begin{proof}
With no information on which $\mu_i$ each $y_i$ conditions on, we introduce an unobserved indicator variable $z_i$ that is $k$ if $y_i | z_i \sim N(\tau_k, \sigma^2)$. Note that
 \[z\sim Multinomial(M, \pi).\]
 
The full likelihood function then becomes   
	     \begin{equation}
	     	\label{lik-npeb}
			\begin{split}
	     	L(\y ~|~  \z, \pi)  &= p\func{ \y~|~\z } p\func{ \z~|~ \pi } \\
	     	&= \prod_{i=1}^n \prod_{j=1}^M  \left( p\func{y_i~|~ \tau_j} \cdot \pi_j \right)^{\bm 1_{\{z_i = j\}}}
	   		\end{split}
	     \end{equation}

 and thus the log-likelihood  
 \begin{equation}
 	\label{loglik-npeb}
	\begin{split}
 	\ell(\y ~|~ \z, \bm \pi) = \sum_{i=1}^n \sum_{j=1}^M  {\bm 1_{\{z_i = j\}}} \left (\log p\func{y_i~|~\tau_j} + \log \pi_j \right).
\end{split}
 \end{equation}

In the E-step, we take expected value of the above likelihood function over $\z$ conditional on $\y$ and  $\piold$, i.e. $\Eold$ refers to averaging $z$ over the distribution $\Prob(\z ~|~ \piold, \y)$. Given the full log-likehood (\ref{loglik-npeb}), the expected log posterior density is
\begin{equation}
	\label{epost}
	\Eold_{Z|\piold, y} \ell (  \z, \pi ~|~ \y) =  \sum_{i=1}^n \sum_{j=1}^M  \Prob\func{Z_i = j | y_i , \piold} \left (\log p\func{y_i~|~\tau_j} + \log \pi_j \right).
\end{equation}

For the M-step, we maximize (\ref{epost}) over $\bm \pi$.  Let 
\begin{equation}
	\label{pi-def}
	\piold_{j,i} = \Prob(Z_i = j ~|~ Y_i = y_i, \piold) = \frac{p(y_i ~|~ \tau_j) \piold_j}{\sum_{m=1}^M f(y_i~|~ \tau_m) \piold_m}
\end{equation}
 Since the only unknowns in (\ref{pi-def}) are $\pi_j$'s with the constraint of $\sum \pi_j =1$, it is a constrained optimization problem
\begin{equation}
	\label{pi-update}
	\begin{split}
		&\argmax_\pi \sum_{i=1}^n \sum_{j=1}^M  \Prob\func{Z_i = j | y_i , \piold} \left (\log p\func{y_i~|~\tau_j} + \log \pi_j \right) \\
		&\text{subject to } \sum \pi_j =1
	\end{split}
\end{equation}
By the method of Lagrange multipliers, we find the maximizer being 
\begin{equation}
	\label{pij-upate}
	\pinew_j = \frac{\sum_{i=1}^{n}\piold_{j,i}}{\sum_{k=1}^M \sum_{i=1}^{n}\piold_{k,i}} = \frac{1}{n}\sum_{i=1}^{n}\piold_{j,i}.
\end{equation}
 \end{proof}

%\begin{proof}[Proof of Theorem \ref{Main-thm:dnp-em}]
\begin{proof}[Proof of Theorem 1]
The algorithm converges to the maximizer of the likelihood, see Wu (1983). The set of sub-stochastic distributions is compact, and clearly the maximizer would be a properly stochastic distribution with no mass escaping to infinite. The uniqueness of the Gaussian deconvolution ensures the consistency of the distribution of $\bpi$. Finally, if $\bpi_M^0$ is the  population discretized MLE on the grid, then clearly $\bpi^0_M \rightarrow \bpi^0$ as $M\rightarrow \infty$.

%The general consistency of the nonparametric maximum likelihood estimator was established in \cite{kiefer1956consistency}. Our setup is a special case of Example 1 in \cite{kiefer1956consistency}, i.e., for any $p(y | \mu)$ in the exponential family, assuming if $\pi_1$ and $\pi_2$ are two different distributions on $\mu$, then for at least one $y$, $m(y | \pi_1) \neq m(y | \pi_2)$ where $m(y|\pi)$ is the marginal density of $y$, i.e. $m(y|\pi) = \int p(y|\mu) \pi(\mu) d\mu$.
\end{proof}

It is worth mentioning that the likelihood has a convex geometry. The functional of interest is
\begin{equation}
L(\pi) =  \sum_{i=1}^n \Big\{-\log \int p(y_i|\mu_i) \pi(\mu_i) d\mu_i \Big\},
\end{equation}
which is a sum of $-\log(\cdot)$ of a linear functional $\int p(y_i|\mu_i) \pi(\mu_i) d\mu_i$. Since $-\log(\cdot)$ is convex and non-decreasing, $L(\pi)$ is convex if the space of $\pi$ is convex.
This is also the case for the discretized version, 
\begin{equation}
L(\pi) =  \sum_{i=1}^n \Big\{-\log \sum_j p(y_i|\tau_j) \pi_j \Big\}.
\end{equation}

%%%%%%%%%%%%%%%%%%%%%%%%%%%%%%%%%%%%%%%%%%%%%%%%%%%%%%%%%%%%
%\subsection{Proof of Proposition \ref{Main-prop1}}
\subsection{Proof of Proposition 1}

Let a decision vector be $\a=(a_1,\cdots,a_m)$ where $a_i=1$ if we reject the $i$-th hypothesis and $a_i=0$ otherwise. A false discovery can be expressed as $a_i \I_{\mu_i= 0}$ where $\I$ is an indicator function, and similarly a false non-discovery as $(1-a_i)\I_{\mu_i\neq0}$. 

An optimal testing procedure can be constructed to minimize the objective function $\E(T)$, subject to $R=k$ for a positive integer number $k$. That is, given the number of total discoveries, we want to minimize the averaged number of false non-discoveries. 
Correspondingly, the objective function can be written as
\begin{equation}\label{opt}
\min_{(a_1,\cdots,a_m)}\E\Big[\sum_{i=1}^m(1-a_i)\I_{\mu_i\neq0}\Big]
\quad \text{s.t.} \quad \sum_{i=1}^ma_i=k.
\end{equation}

 The expectation in (\ref{opt}) is over the distribution of $\y$ and the prior distribution of $\bmu$. It is equivalent to minimize the expectation of loss function conditional on $\y$, that is, minimizing
\begin{eqnarray}\label{eq20}
&&\min_{(a_1,\cdots,a_m)} L(\a) = \sum_{i=1}^m\Big[(1-a_i)P(\mu_i\neq0|\y)\Big]\\
\text{subject to} && \sum_{i=1}^m a_i=k. \nonumber 
\end{eqnarray}

After some algebra, (\ref{eq20}) can be re-arranged as 
\begin{equation*}
L(\a)=\sum_{i=1}^m(1-P(\mu_i=0|\y))+\sum_{i=1}^ma_i[P(\mu_i=0|\y)-1]. 
\end{equation*}
Note that $L(\a)$ is increasing when the second term is increasing. Therefore, to minimize $L(\a)$ subject to $\sum_{i=1}^ma_i=k$, for the smallest $k$ values of $P(\mu_i=0|\y)$, we set the corresponding $a_i=1$, and the rest of the $a_i$'s are set as 0. The proof is now complete. 

%%%%%%%%%%%%%%%%%%%%%%%%%%%%%%%%%%%%%%%%%%%%%%%%%%%%%%%%%%%%
%%%%%%%%%%%%%%%%%%%%%%%%%%%%%%%%%%%%%%%%%%%%%%%%%%%%%%%%%%%%
\section{Simulation tables}
\label{app:sim-tbl}
%%%%%%%%%%%%%%%%%%%%%%%%%%%%%%%%%%%%%%%%%%%%%%%%%%%%%%%%%%%%
%%%%%%%%%%%%%%%%%%%%%%%%%%%%%%%%%%%%%%%%%%%%%%%%%%%%%%%%%%%%

We provide all the information of Figure \ref{plot:het-mse} to \ref{plot:het-emp-power-w95} in the table format for detailed comparison. In each row, the best performer is set in bold type. Note that we show the bias of sparsity estimation and we consider the method yielding the smallest absolute value as the best performer. 

\foreach \n in {1,1.5,...,2.5}{
\def\uu{\n}
\hettbl{MSE}{\uu}
\hettbl{MSE_ratio}{\uu}
\hettbl{MSE_mode}{\uu}
\hettbl{MSE_mode_ratio}{\uu}
\hettbl{w_diff}{\uu}
\hettbl{ci_coverage}{\uu}
\hettbl{ci_len}{\uu}
\hettbl{emp_fdr_V1}{\n}
\hettbl{emp_fdr_V2}{\n}
\hettbl{emp_power_V1}{\n}
\hettbl{emp_power_V2}{\n}
\clearpage
}

%%%%%%%%%%%%%%%%%%%%%%%%%%%%%%%%%%%%%%%%%%%%%%%%%%%%%%%%%%%%

\bibliographystyle{plainnat}
\bibliography{biblio}

%%%%%%%%%%%%%%%%%%%%%%%%%%%%%%%%%%%%%%%%%%%%%%%%%%%%%%%%%%%%